\pdfoutput=1
\documentclass[12pt]{article}
\usepackage{amsmath}
\usepackage{natbib}
\usepackage{microtype}
\usepackage{graphicx}
\usepackage{subfig}
\usepackage{booktabs}
\usepackage[table]{xcolor} 
\usepackage{colortbl} 
\usepackage{bm}
\usepackage{pgf, tikz} 
\usetikzlibrary{arrows,automata,fit}
\usetikzlibrary{shapes,snakes}
\usepackage{bbm} 
\usepackage{amsfonts}
\usepackage{amsthm}
\usepackage{bm}
\usepackage{verbatim}
\newtheorem{definition}{Definition}
\newtheorem{proposition}{Proposition}

\newtheorem{corollary}{Corollary}
\newtheorem{lemma}{Lemma}
\usepackage{algorithm}
\usepackage{algorithmic}

\newcommand{\xx}{1}
\newcommand{\yy}{1}
\newcommand{\sage}[2]{\tikz{\node[shape=circle,draw,inner sep=1pt,minimum width = 0.6cm, fill=#1]{$v_{#2}$};}} 
 
\newcommand{\stages}[2]{\tikz{\node[shape=circle,draw,inner sep=1pt,fill=#1,minimum size=0.5cm]{${#2}$};}} 
\newcommand{\stag}[2]{\tikz{\node[shape=circle,draw,inner sep=1pt,fill=#1,minimum size=1cm]{${#2}$};}} 
\newcommand{\leaf}{\tikz{\node[shape=circle,draw,inner sep=1.5pt,fill=white] {};}}

\newcommand\independent{\protect\mathpalette{\protect\independenT}{\perp}}
\def\independenT#1#2{\mathrel{\rlap{$#1#2$}\mkern2mu{#1#2}}}

\usepackage{hyperref}

\newcommand{\blind}{1}

\begin{document}

\def\spacingset#1{\renewcommand{\baselinestretch}%
{#1}\small\normalsize} \spacingset{1}


\if1\blind
{
 \title{\bf Staged Trees and Asymmetry-Labeled DAGs}
  \author{Gherardo Varando\\
Image Processing Laboratory, Universitat de València,  València, Spain\\ 
and\\
Federico Carli  \\
Dipartimento di Matematica, Universit\`{a} degli Studi di Genova, Genova, Italy\\
and\\
Manuele Leonelli \\
School of Science and Technology, IE University, Madrid, Spain}
  \maketitle
} \fi

\if0\blind
{
  \bigskip
  \bigskip
  \bigskip
  \begin{center}
    {\LARGE\bf Title}
\end{center}
  \medskip
} \fi

\bigskip
\begin{abstract}
Bayesian networks are a widely-used class of probabilistic graphical models capable of representing symmetric conditional independence between variables of interest using the topology of the underlying graph. For categorical variables, they can be seen as a special case of the much more general class of models called staged trees, which can represent any type of non-symmetric conditional independence. 
Here we formalize the relationship between these two models and introduce a minimal Bayesian network representation of the staged tree, which can be used to read conditional independences in an intutitive way. A new labeled graph termed asymmetry-labeled directed acyclic graph is defined, whose edges are labeled to denote the type of dependence existing between any two random variables.
We also present  a novel algorithm to learn staged 
trees which only enforces a specific subset 
of non-symmetric independences. 
Various datasets are used to illustrate the methodology, highlighting the need to construct models which more flexibly encode and represent non-symmetric structures.
\end{abstract}

\noindent%

\spacingset{1.45} 

\section{Introduction}
Probabilistic graphical models give an intuitive and efficient representation of the relationships existing between random variables of interest. Bayesian networks (BNs) \citep[see e.g.][]{Darwiche2009} are the most commonly used graphical model and have been applied in a variety of real-world applications.
One of the main limitations of BNs is that they can only represent symmetric conditional independences, which in practice can be too restrictive.

For this reason, \citet{Boutilier1996} introduced the notion of \emph{context-specific} independence, meaning that independences hold only for specific values, or \textit{contexts}, of the conditioning variables. Extensions of BNs encoding context-specific independences are usually defined by associating a tree representation to each vertex of the network \citep{Cano2012,Friedman1996,Talvitie2019}, by labeling the edges \citep{Pensar2015,Hyttinen2018}, or by using some alternative approach \citep{Chickering1997, Geiger1996,Poole2003}. In recent years there has been a growing interest in formalizing context-specific independence \citep{Corander2019,Shen2020} and in generalizing other graphical models with non-symmetric dependencies \citep{Nyman2016,Pensar2017}.

With the exception of \citet{Jaeger2006} and \citet{Pensar2015}, BNs embellished with context-specific independence lose their intuitiveness since all the model information cannot be succinctly represented in a unique graph. Staged trees \citep{Smith2008,Collazo2018} are probabilistic graphical models that, starting from an event tree, represent non-symmetric conditional independence statements via a coloring of its vertices. Coloring has recently been found to provide a valuable embellishment to other graphical models \citep{Hojsgaard2008, Massam2018}.

As demonstrated by \citet{Smith2008} and \citet{Duarte2020}, every BN can be represented as a staged tree. However, the class of staged tree models is much more general and can represent not only symmetric, but also context-specific, partial and local independences \citep{Pensar2016}. 
 Furthermore, a wide array of methods to efficiently investigate real-world applications have been introduced for staged trees, including user-friendly software \citep{Carli2020}, inferential routines \citep{Gorgen2015}, structural learning \citep{Freeman2011}, dealing with missing data \citep{Barclay2014}, causal reasoning \citep{Thwaites2010} and identification of equivalence classes \citep{Gorgen2018}, to name a few. Such techniques are in general not available for other graphical models embedding non-symmetric independences, thus making staged trees a viable as well as efficient option for applied analyses.

Our first contribution is a deeper study of the relationship between BNs and staged trees. We introduce a minimal BN representation of a staged tree which embeds all its symmetric conditional independences. Importantly, this allows us to introduce a criterion to identify all symmetric conditional independences implied by the model, which has proven to be a very challenging task \citep{Thwaites2015}.

Reading non-symmetric independences directly from the staged tree is even more challenging. Our second contribution is a novel definition of classes of dependence among variables and the introduction of methods to identify the appropriate class from the staged tree. The presence or absence of edges in a BN encodes either (conditionally) full dependence or independence between two variables. However, the flexibility of the staged tree enables us to model and consequently identify intermediate relationship between variables, namely context-specific, partial or local \citep{Pensar2016}. 

As a result, our third contribution is the definition of a new class of directed acyclic graphs (DAGs), termed \emph{asymmetry-labeled DAGs} (ALDAGs), by coloring edges according to the type of relationship existing between the corresponding variables. Learning algorithms for ALDAGs, which use any structural learning algorithm for staged trees \citep[see e.g. those included in the R package \texttt{stagedtrees},][]{Carli2020}, are discussed below and applied to various datasets. Our fourth contribution is the definition of a new visualization of dependence, called the \textit{dependence subtree}, which shows how a variable is related to only those that have a direct effect on it, namely its parents in the associated ALDAG. The use of such a tool is showcased in our data applications below.

Structural learning of generic staged trees is hard, due to the explosion of the model search space as the number of variables increases \citep[see e.g.][]{duarte2021representation}. For this reason, recent research has focused on sub-classes of staged tree models:  \citet{Carli2020a} defined
naive staged trees which have the same number of parameters of a naive BN over the same variables; \citet{leonelli2022structural} considered simple staged trees which have a constrained type
of partitioning of the vertices; \citet{leonelli2022highly} introduced $k$-parents staged trees which limit the number of variables that can have a direct influence on another; \citet{duarte2021representation} defined CStrees which only embed
symmetric and context-specific types of independence. Our last contribution is the introduction of a novel
algorithm, called \emph{context-specific backward hill-climbing (CSBHC)}, to learn staged trees whose staging is restricted. In particular, the proposed algorithm learns staged trees whose corresponding ALDAGs have a restricted subset of labels, those corresponding to conditional 
independences of the  context-specific type only. 



\section{Bayesian Networks and Conditional Independence}
\label{sec:bn}
 Let $G=([p],F)$ be a directed acyclic graph (DAG) with vertex set $[p]=\{1,\dots,p\}$ and edge set $F$. Let $\bm{X}=(X_i)_{i\in[p]}$ be categorical random variables with joint mass function $P$ and sample space $\mathbb{X}=\times_{i\in[p]}\mathbb{X}_i$. For $A\subset [p]$, we let $\bm{X}_A=(X_i)_{i\in A}$ and $\bm{x}_A=(x_i)_{i\in A}$ where $\bm{x}_A\in\mathbb{X}_A=\times_{i\in A}\mathbb{X}_i$. We say that $P$ is Markov to $G$ if, for $\bm{x}\in\mathbb{X}$, 
\[
P(\bm{x})=\prod_{k\in[p]}P(x_k | \bm{x}_{\Pi_k}),
\]
where $\Pi_k$ is the parent set of $k$ in $G$ and $P(x_k | \bm{x}_{\Pi_k})$ is a shorthand for $P(X_k=x_k |\bm{X}_{\Pi_k} = \bm{x}_{\Pi_k})$. The ordered Markov condition implies conditional independences of the form
\begin{equation}
\label{ci}
X_i \independent \bm{X}_{[i-1]}\,|\, \bm{X}_{\Pi_i},
\end{equation}
which are equivalent to 
\begin{equation}
\label{ci2}
P(x_i|\bm{x}_{[i-1]\setminus \Pi_i}, \bm{x}_{\Pi_i})= P(x_i|\bm{x}_{\Pi_i}), \hspace{0.5cm} \mbox{for all }  \bm{x}\in\mathbb{X}.
\end{equation}
\begin{definition}
	The \emph{Bayesian network} model (associated to $G$) is 
\[
\mathcal{M}_G = \{P\in\Delta_{|\mathbb{X}|-1}\,|\, P \mbox{ is Markov to } G\},
\]
where $\Delta_{|\mathbb{X}|-1}$ is the ($|\mathbb{X}|-1$)-dimensional probability simplex.
\end{definition}

It is customary to label the vertices of a BN so to respect the topological order of $G$, i.e. a linear ordering of $[p]$ for which only pairs $(i,j)$ where $i$ appears before $j$ in the order can be in the edge set. Of course, there can be multiple permutations of $[p]$ which respect the topological order. Henceforth, we assume that the natural ordering of the positive  integers $[p]$ respects the topological order of the BN.

\begin{figure}
\centering
\begin{tikzpicture}
\renewcommand{\xx}{2}
\renewcommand{\yy}{1.5}
\node (1) at (1*\xx,0*\yy){\stages{white}{C}};
\node (2) at (1*\xx,1*\yy){\stages{white}{S}};
\node (3) at (0*\xx,0*\yy){\stages{white}{G}};
\node (4) at (2*\xx,0*\yy){\stages{white}{A}};
\draw[->, line width = 1.1pt] (1) -- (2);
\draw[->, line width = 1.1pt] (1) -- (3);
\draw[->, line width = 1.1pt] (1) -- (4);
\draw[->, line width = 1.1pt] (3) -- (2);
\draw[->, line width = 1.1pt] (2) -- (4);
\end{tikzpicture}
\caption{Learned BN for the \texttt{Titanic} dataset. \label{fig:bn}}
\end{figure}
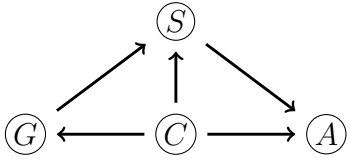

To illustrate our methodology, we use throughout the paper the \texttt{Titanic} dataset \citep{Dawson1995} which provides information on the fate of the Titanic passengers and available from the \texttt{datasets} package bundled in R. \texttt{Titanic} includes four categorical variables:  \texttt{Class} (C) has four levels whilst \texttt{Gender} (G), \texttt{Survived} (S) and  \texttt{Age} (A)  are binary. The BN learned using the hill-climbing algorithm implemented in the R package \texttt{bnlearn} \citep{Scutari2010} is reported in Figure \ref{fig:bn} and embeds the conditional independence $
\texttt{Age}\independent \texttt{Gender} ~ | ~ \texttt{Class}, \texttt{Survived}$ only.
Only one topological order of the variables exists  and is henceforth used: C, G, S, A.

\subsection{Non-Symmetric Conditional Independence}

BNs have the capability of expressing only symmetric conditional independences of the form in (\ref{ci}) and (\ref{ci2}). The most common non-symmetric extension of conditional independence is the so-called context-specific independence which is often represented associating a tree to each vertex of a BN \citep{Boutilier1996}.  Let $A$, $B$ and $C$ be three disjoint subsets of $[p]$. We say that $\bm{X}_A$ is context-specific independent of $\bm{X}_B$ given context $\bm{x}_C\in\mathbb{X}_C$ if 
\begin{equation}
\label{eq:csi}
P(\bm{x}_A|\bm{x}_B,\bm{x}_C)=P(\bm{x}_A|\bm{x}_C)
\end{equation}
holds for all $(\bm{x}_A,\bm{x}_B)\in\mathbb{X}_{A\cup B}$ and write $\bm{X}_A\independent \bm{X}_B|\bm{x}_C$. The condition in (\ref{eq:csi}) reduces to standard conditional independence in (\ref{ci}) if it holds for all $\bm{x}_C\in\mathbb{X}_C$.

\citet{Pensar2016} introduced a more general definition of non-symmetric conditional independence called \emph{partial conditional independence}. We say that $\bm{X}_A$ is partially conditionally independent of $\bm{X}_B$ in the domain $\mathcal{D}_B\subseteq \mathbb{X}_B$ given context $\bm{X}_C=\bm{x}_C$ if
\begin{equation}
\label{eq:pci}
P(\bm{x}_A|\bm{x}_B,\bm{x}_C)=P(\bm{x}_A|\tilde{\bm{x}}_B,\bm{x}_C)
\end{equation}
holds for all $(\bm{x}_A,\bm{x}_B),(\bm{x}_A,\tilde{\bm{x}}_B)\in\mathbb{X}_A\times \mathcal{D}_B$ and write $\bm{X}_A\independent \bm{X}_B|\mathcal{D}_B,\bm{x}_C$. Clearly, (\ref{eq:csi}) and $(\ref{eq:pci})$ coincide if $\mathcal{D}_B=\mathbb{X}_B$. Furthermore, the sample space $\mathbb{X}_B$ must contain more than two elements for a non-trivial partial conditional independence to hold.

A final condition is the so called \emph{local conditional independence} and first discussed in \citet{Chickering1997}. For $i\in[p]$ and an $A\subset[p]$ such that $A\cap \{i\}=\emptyset$, local conditional independence expresses equalities of probabilities of the form
\begin{equation}
\label{eq:lci}
P(x_i|\bm{x}_A)= P(x_i|\tilde{\bm{x}}_A)
\end{equation}
for all $x_i\in\mathbb{X}_i$ and two $\bm{x}_A,\tilde{\bm{x}}_A\in\mathbb{X}_A$. Notice that in terms of generality, (\ref{eq:csi}) $\preceq$ (\ref{eq:pci}) $\preceq$ (\ref{eq:lci}). Condition (\ref{eq:lci}) simply states that some conditional probability distributions are identical, where no discernable patterns as in (\ref{eq:csi}) and (\ref{eq:pci}) can be detected.

Differently to any other probabilistic graphical model, the class of staged trees that we review next is able to graphically represent and formally encode any of the types of conditional independences defined in (\ref{ci2})-(\ref{eq:lci}).

\section{Staged Trees}



Differently to BNs, whose graphical representation is a DAG, staged trees visualize conditional independence by means of a colored tree. Let $(V,E)$ be a directed, finite, rooted tree with vertex set $V$, root node $v_0$ and edge set $E$. 
For each $v\in V$, 
let $E(v)=\{(v,w)\in E\}$ be the set of edges emanating
from $v$ and $\mathcal{C}$ be a set of labels. 

An $\bf X$-compatible staged tree 
is a triple $T = (V,E,\theta)$, where $(V,E)$ is a rooted directed tree and:
\begin{enumerate}
    \item $V = {v_0} \cup \bigcup_{i \in [p]} \mathbb{X}_{[i]}$;
		\item For all $v,w\in V$,
$(v,w)\in E$ if and only if $w=\bm{x}_{[i]}\in\mathbb{X}_{[i]}$ and 
			$v = \bm{x}_{[i-1]}$, or $v=v_0$ and $w=x_1$ for some
$x_1\in\mathbb{X}_1$;
\item $\theta:E\rightarrow \mathcal{L}=\mathcal{C}\times \cup_{i\in[p]}\mathbb{X}_i$ is a labelling of the edges such that $\theta(v,\bm{x}_{[i]}) = (\kappa(v), x_i)$ for some 
			function $\kappa: V \to \mathcal{C}$. The function 
			$k$ is called the \textit{colouring} of the staged tree $T$.
\end{enumerate}
	If $\theta(E(v)) = \theta(E(w))$ then $v$ and $w$ are said to be in the same 	\emph{stage}. The equivalence classes induced by  $\theta(E(v))$
form a partition of the internal vertices of the tree  in \emph{stages}.

Points 1 and 2 above construct a rooted tree where each root-to-leaf path, or equivalently each leaf, is associated to an element of the sample space $\mathbb{X}$.  Then a labeling of the edges of such a tree is defined where labels are pairs with one element from a set $\mathcal{C}$ and the other from the sample space $\mathbb{X}_i$ of the corresponding variable $X_i$ in the tree. By construction, $\bf X$-compatible staged trees are such that two vertices can be in the same stage if and only if they correspond to the same sample space.

\begin{figure}
\centering
\scalebox{0.6}{
\begin{tikzpicture}
\renewcommand{\xx}{1.8}
\renewcommand{\yy}{0.75}
\node (v0) at (0*\xx,0*\yy) {\sage{cyan}{0}};
\node (v1) at (1*\xx,-6*\yy) {\sage{cyan}{1}};
\node (v2) at (1*\xx,-2*\yy) {\sage{cyan}{2}};
\node (v3) at (1*\xx,2*\yy) {\sage{red}{3}};
\node (v4) at (1*\xx,6*\yy) {\sage{green}{4}};
\node (v9) at (2*\xx,1*\yy) {\sage{green}{9}};
\node (v10) at (2*\xx,3*\yy) {\sage{cyan}{10}};
\node (v11) at (2*\xx,5*\yy) {\sage{orange}{11}};
\node (v12) at (2*\xx,7*\yy) {\sage{yellow}{12}};
\node (v8) at (2*\xx,-1*\yy) {\sage{yellow}{8}};
\node (v7) at (2*\xx,-3*\yy) {\sage{green}{7}};
\node (v6) at (2*\xx,-5*\yy) {\sage{red}{6}};
\node (v5) at (2*\xx,-7*\yy) {\sage{cyan}{5}};
\node (v21) at (3*\xx,0.5*\yy) {\sage{red}{21}};
\node (v22) at (3*\xx,1.5*\yy) {\sage{yellow}{22}};
\node (v23) at (3*\xx,2.5*\yy) {\sage{orange}{23}};
\node (v24) at (3*\xx,3.5*\yy) {\sage{orange}{24}};
\node (v25) at (3*\xx,4.5*\yy) {\sage{cyan}{25}};
\node (v26) at (3*\xx,5.5*\yy) {\sage{cyan}{26}};
\node (v27) at (3*\xx,6.5*\yy) {\sage{cyan}{27}};
\node (v28) at (3*\xx,7.5*\yy) {\sage{cyan}{28}};
\node (v20) at (3*\xx,-0.5*\yy) {\sage{yellow}{20}};
\node (v19) at (3*\xx,-1.5*\yy) {\sage{cyan}{19}};
\node (v18) at (3*\xx,-2.5*\yy) {\sage{green}{18}};
\node (v17) at (3*\xx,-3.5*\yy) {\sage{cyan}{17}};
\node (v16) at (3*\xx,-4.5*\yy) {\sage{cyan}{16}};
\node (v15) at (3*\xx,-5.5*\yy) {\sage{cyan}{15}};
\node (v14) at (3*\xx,-6.5*\yy) {\sage{red}{14}};
\node (v13) at (3*\xx,-7.5*\yy) {\sage{cyan}{13}};
\node (v45) at (4*\xx,0.35*\yy) {\leaf};
\node (v46) at (4*\xx,0.65*\yy) {\leaf};
\node (v47) at (4*\xx,1.35*\yy) {\leaf};
\node (v48) at (4*\xx,1.65*\yy) {\leaf};
\node (v49) at (4*\xx,2.35*\yy) {\leaf};
\node (v50) at (4*\xx,2.65*\yy) {\leaf};
\node (v51) at (4*\xx,3.35*\yy) {\leaf};
\node (v52) at (4*\xx,3.65*\yy) {\leaf};
\node (v53) at (4*\xx,4.35*\yy) {\leaf};
\node (v54) at (4*\xx,4.65*\yy) {\leaf};
\node (v55) at (4*\xx,5.35*\yy) {\leaf};
\node (v56) at (4*\xx,5.65*\yy) {\leaf};
\node (v57) at (4*\xx,6.35*\yy) {\leaf};
\node (v58) at (4*\xx,6.65*\yy) {\leaf};
\node (v59) at (4*\xx,7.35*\yy) {\leaf};
\node (v60) at (4*\xx,7.65*\yy) {\leaf};
\node (v44) at (4*\xx,-0.35*\yy) {\leaf};
\node (v43) at (4*\xx,-0.65*\yy) {\leaf};
\node (v42) at (4*\xx,-1.35*\yy) {\leaf};
\node (v41) at (4*\xx,-1.65*\yy) {\leaf};
\node (v40) at (4*\xx,-2.35*\yy) {\leaf};
\node (v39) at (4*\xx,-2.65*\yy) {\leaf};
\node (v38) at (4*\xx,-3.35*\yy) {\leaf};
\node (v37) at (4*\xx,-3.65*\yy) {\leaf};
\node (v36) at (4*\xx,-4.35*\yy) {\leaf};
\node (v35) at (4*\xx,-4.65*\yy) {\leaf};
\node (v34) at (4*\xx,-5.35*\yy) {\leaf};
\node (v33) at (4*\xx,-5.65*\yy) {\leaf};
\node (v32) at (4*\xx,-6.35*\yy) {\leaf};
\node (v31) at (4*\xx,-6.65*\yy) {\leaf};
\node (v30) at (4*\xx,-7.35*\yy) {\leaf};
\node (v29) at (4*\xx,-7.65*\yy) {\leaf};
\draw[->] (v0) -- node [below, sloped] {\tiny{1st}} (v1);
\draw[->] (v0) -- node [below, sloped] {\tiny{2nd}} (v2);
\draw[->] (v0) --  node [above, sloped] {\tiny{3rd}} (v3);
\draw[->] (v0) --  node [above, sloped] {\tiny{Crew}} (v4);
\draw[->] (v1) --  node [below, sloped] {\tiny{Male}} (v5);
\draw[->] (v1) --  node [above, sloped] {\tiny{Female}} (v6);
\draw[->] (v2) --  node [below, sloped] {\tiny{Male}} (v7);
\draw[->] (v2) --  node [above, sloped] {\tiny{Female}} (v8);
\draw[->] (v3) --  node [below, sloped] {\tiny{Male}} (v9);
\draw[->] (v3) --  node [above, sloped] {\tiny{Female}} (v10);
\draw[->] (v4) --  node [below, sloped] {\tiny{Male}} (v11);
\draw[->] (v4) --  node [above, sloped] {\tiny{Female}} (v12);
\draw[->] (v5) --  node [below, sloped] {\tiny{No}} (v13);
\draw[->] (v5) --  node [above, sloped] {\tiny{Yes}} (v14);
\draw[->] (v6) --  node [below, sloped] {\tiny{No}} (v15);
\draw[->] (v6) --  node [above, sloped] {\tiny{Yes}} (v16);
\draw[->] (v7) --  node [below, sloped] {\tiny{No}} (v17);
\draw[->] (v7) --  node [above, sloped] {\tiny{Yes}} (v18);
\draw[->] (v8) --  node [below, sloped] {\tiny{No}} (v19);
\draw[->] (v8) --  node [above, sloped] {\tiny{Yes}} (v20);
\draw[->] (v9) --  node [below, sloped] {\tiny{No}} (v21);
\draw[->] (v9) --  node [above, sloped] {\tiny{Yes}} (v22);
\draw[->] (v10) --  node [below, sloped] {\tiny{No}} (v23);
\draw[->] (v10) --  node [above, sloped] {\tiny{Yes}} (v24);
\draw[->] (v11) --  node [below, sloped] {\tiny{No}} (v25);
\draw[->] (v11) --  node [above, sloped] {\tiny{Yes}} (v26);
\draw[->] (v12) --  node [below, sloped] {\tiny{No}} (v27);
\draw[->] (v12) --  node [above, sloped] {\tiny{Yes}} (v28);
\draw[->] (v13) --  node [below, sloped] {\tiny{Child}} (v29);
\draw[->] (v13) --  node [above, sloped] {\tiny{Adult}} (v30);
\draw[->] (v14) --  node [below, sloped] {\tiny{Child}} (v31);
\draw[->] (v14) --  node [above, sloped] {\tiny{Adult}} (v32);
\draw[->] (v15) --  node [below, sloped] {\tiny{Child}} (v33);
\draw[->] (v15) --  node [above, sloped] {\tiny{Adult}} (v34);
\draw[->] (v16) --  node [below, sloped] {\tiny{Child}} (v35);
\draw[->] (v16) --  node [above, sloped] {\tiny{Adult}} (v36);
\draw[->] (v17) --  node [below, sloped] {\tiny{Child}} (v37);
\draw[->] (v17) --  node [above, sloped] {\tiny{Adult}} (v38);
\draw[->] (v18) --  node [below, sloped] {\tiny{Child}} (v39);
\draw[->] (v18) --  node [above, sloped] {\tiny{Adult}} (v40);
\draw[->] (v19) --  node [below, sloped] {\tiny{Child}} (v41);
\draw[->] (v19) --  node [above, sloped] {\tiny{Adult}} (v42);
\draw[->] (v20) --  node [below, sloped] {\tiny{Child}} (v43);
\draw[->] (v20) --  node [above, sloped] {\tiny{Adult}} (v44);
\draw[->] (v21) --  node [below, sloped] {\tiny{Child}} (v45);
\draw[->] (v21) --  node [above, sloped] {\tiny{Adult}} (v46);
\draw[->] (v22) --  node [below, sloped] {\tiny{Child}} (v47);
\draw[->] (v22) --  node [above, sloped] {\tiny{Adult}} (v48);
\draw[->] (v23) --  node [below, sloped] {\tiny{Child}} (v49);
\draw[->] (v23) --  node [above, sloped] {\tiny{Adult}} (v50);
\draw[->] (v24) --  node [below, sloped] {\tiny{Child}} (v51);
\draw[->] (v24) --  node [above, sloped] {\tiny{Adult}} (v52);
\draw[->] (v25) --  node [below, sloped] {\tiny{Child}} (v53);
\draw[->] (v25) --  node [above, sloped] {\tiny{Adult}} (v54);
\draw[->] (v26) --  node [below, sloped] {\tiny{Child}} (v55);
\draw[->] (v26) --  node [above, sloped] {\tiny{Adult}} (v56);
\draw[->] (v27) --  node [below, sloped] {\tiny{Child}} (v57);
\draw[->] (v27) --  node [above, sloped] {\tiny{Adult}} (v58);
\draw[->] (v28) --  node [below, sloped] {\tiny{Child}} (v59);
\draw[->] (v28) --  node [above, sloped] {\tiny{Adult}} (v60);
\end{tikzpicture}
}

\caption{A staged tree compatible with $(\texttt{Class}, \texttt{Gender}, \texttt{Survived}, \texttt{Age})$, learned over the \texttt{Titanic} dataset. \label{fig:staged1}}
\end{figure}
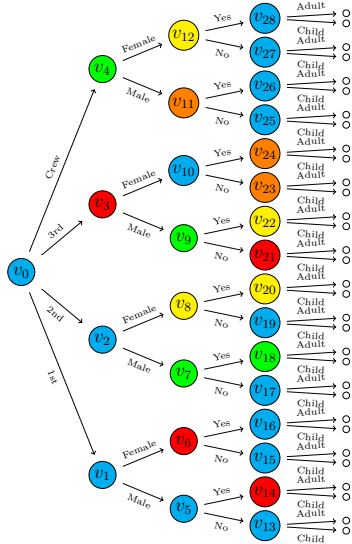

Figure \ref{fig:staged1} reports an example of an $\bf X$-compatible staged tree model 
for the Titanic dataset
learned with the R package \texttt{stagedtrees}. 
The \textit{coloring} given by the function $\kappa$ is shown in the vertices and
each edge $(\cdot , (x_1, \ldots, x_{i}))$ is labeled with $x_{i}$.
The edge labeling $\theta$ can be read from the graph combining the text label and the 
color of the emanating vertex. For example, $ \theta(v_1, v_6) \neq \theta(v_1, v_5) 
= \theta(v_3,v_{10}) \neq \theta(v_2, v_{7}) \neq \theta(v_5, v_{14})$.  This representation of the labeling $\theta$  over vertices is equivalent to that over edges, whilst being more interpretable, and is henceforth used.
There are 29 internal vertices and the staging is $\{v_0\}$, $\{v_1,v_2\}$, $\{v_3\}$,
$\{v_4\}$, $\{v_5,v_{10}\}$,  $\{v_6\}$, $\{v_7,v_9\}$, $\{v_8,v_{12}\}$,
$\{v_{11}\}$, $\{v_{13},v_{15},v_{16},v_{17},v_{19},v_{25},v_{26},
v_{27},v_{28}\}$, $\{v_{14},v_{21},v_{22}\}$, $\{v_{18}\}$, $\{v_{20},v_{22}\}$
and $\{v_{23},v_{24}\}$.

The parameter space associated to an $\bf X$-compatible staged tree $T = (V, E, \theta)$ 
with 
labeling $\theta:E\rightarrow \mathcal{L}$ 
is defined as
\begin{equation}
\label{eq:parameter}
	\Theta_T=\Big\{\bm{y}\in\mathbb{R}^{\vert \theta(E)\vert} \;\vert \; \forall ~ e\in E, y_{\theta(e)}\in (0,1)\textnormal{ and }\sum_{e\in E(v)}y_{\theta(e)}=1\Big\}.
\end{equation}
Equation~(\ref{eq:parameter}) defines a class of probability mass functions 
over the edges emanating from any internal vertex coinciding with conditional distributions  $P(x_i \vert \bm{x}_{[i-1]})$, $\bm{x}\in\mathbb{X}$ and $i\in[p]$.

Let $\bm{l}_{T}$ denote the leaves of a staged tree $T$. Given a vertex $v\in V$, there is a unique path in $T$ from the root $v_0$ to $v$, denoted as $\lambda(v)$. 	The \emph{depth} of a vertex $v\in V$ equals the number of edges in $\lambda(v)$. For any path $\lambda$ in $T$, let $E(\lambda)=\{e\in E: e\in \lambda\}$ denote the set of edges in the path $\lambda$.

\begin{definition}
	The \emph{staged tree model} $\mathcal{M}_{T}$ associated to the $\bf X$-compatible staged 
	tree $(V,E,\theta)$ is the image of the map
\begin{equation}
\label{eq:model}
\begin{array}{llll}
\phi_T & : &\Theta_T &\to \Delta_{\vert\bm{l}_T\vert - 1}^{\circ} \\
 &  & \bm{y} &\mapsto \Big(\prod_{e\in E(\lambda(l))}y_{\theta(e)}\Big)_{l\in \bm{l}_T}
\end{array}
\end{equation}
\end{definition}

Therefore, staged tree models are such that atomic probabilities are equal to the product of the edge labels in root-to-leaf paths and coincide with the usual factorization of mass functions via recursive conditioning.

Conditional independence is formally modeled and represented in staged trees via the labeling $\theta$. As an illustration, consider the staged tree in Figure \ref{fig:staged1} for the \texttt{Titanic} dataset. The fact that $v_1$ and $v_2$ are in the same stage represents the partial independence $\texttt{Gender}\independent \texttt{Class}|\{\texttt{1st},\texttt{2nd}\}$. Considering vertices at depth two, the green and yellow staging again represents partial conditional independences. More interesting is the blue staging of the vertices $v_5$ and $v_{10}$ which implies 
$
P(S = s ~ | ~ \texttt{Female}, \texttt{3rd}) = P(S = s ~ | ~ \texttt{Male}, \texttt{1st}),$ 
i.e. the probability of survival for females travelling in third class is the same as that of males travelling in first class. Such a statement is a generic local conditional independence. Considering the last level, we can notice a very non-symmetric staging structure. As an illustration, consider the top four vertices $v_{25}$, $v_{26}$, $v_{27}$ and $v_{28}$ belonging to the same stage. This implies the context-specific independence
$
\texttt{Age}\independent \texttt{Survived}, \texttt{Gender} ~ | ~ \texttt{Class} = \texttt{Crew}.
$
 The staged tree in Figure \ref{fig:staged1}, embedding the above non-symmetric conditional independences, gives a better representation of the data than the BN in Figure \ref{fig:bn}. Indeed, the BIC of the staged tree can be computed as 10440.39, whilst the one of the BN is larger and equal to 10502.28 \citep[see][for a discussion of using the BIC for staged trees]{Gorgen2020}.

This example illustrates the capability of staged trees to graphically represent any type of non-symmetric conditional independence. Although such independences can be read directly from the tree via visual inspection, it becomes challenging to detect them as the size of the tree increases.  Below we formalize how to assess the type of conditional independence existing between pairs of random variables.

\section{Staged Trees and Bayesian Networks}

Although the relationship between BNs and staged trees was already formalized 
by \citet{Smith2008}, we introduce here an implementable routine to transform a DAG to its equivalent staged tree.

Assume  $\bm{X}$ is topologically ordered with respect to a DAG
$G$
and consider an $\bf X$-compatible staged tree with vertex set $V$,  
edge set $E$ and labeling $\theta$ defined via the 
coloring 
$\kappa(\bm{x}_{[i]} ) = \bm{x}_{\Pi_{i}}$ of the vertices. 
The staged tree $T_G$, with vertex set $V$, edge set $E$ and labeling $\theta$
so constructed, is called \emph{the staged tree model of $G$}. 
Importantly,
$\mathcal{M}_G= \mathcal{M}_{T_G}$, i.e. the two models are exactly the same,
since they entail exactly the same factorization of the joint
probability. The staging of $T_G$ represents the
Markov conditions associated to the graph $G$. 

\begin{figure}
\centering
\scalebox{0.65}{
\begin{tikzpicture}
\renewcommand{\xx}{1.8}
\renewcommand{\yy}{0.75}
\node (v0) at (0*\xx,0*\yy) {\sage{cyan}{0}};
\node (v1) at (1*\xx,-6*\yy) {\sage{cyan}{1}};
\node (v2) at (1*\xx,-2*\yy) {\sage{red}{2}};
\node (v3) at (1*\xx,2*\yy) {\sage{green}{3}};
\node (v4) at (1*\xx,6*\yy) {\sage{yellow}{4}};
\node (v9) at (2*\xx,1*\yy) {\sage{blue}{9}};
\node (v10) at (2*\xx,3*\yy) {\sage{magenta}{10}};
\node (v11) at (2*\xx,5*\yy) {\sage{orange}{11}};
\node (v12) at (2*\xx,7*\yy) {\sage{purple}{12}};
\node (v8) at (2*\xx,-1*\yy) {\sage{yellow}{8}};
\node (v7) at (2*\xx,-3*\yy) {\sage{green}{7}};
\node (v6) at (2*\xx,-5*\yy) {\sage{red}{6}};
\node (v5) at (2*\xx,-7*\yy) {\sage{cyan}{5}};
\node (v21) at (3*\xx,0.5*\yy) {\sage{blue}{21}};
\node (v22) at (3*\xx,1.5*\yy) {\sage{purple}{22}};
\node (v23) at (3*\xx,2.5*\yy) {\sage{blue}{23}};
\node (v24) at (3*\xx,3.5*\yy) {\sage{purple}{24}};
\node (v25) at (3*\xx,4.5*\yy) {\sage{orange}{25}};
\node (v26) at (3*\xx,5.5*\yy) {\sage{magenta}{26}};
\node (v27) at (3*\xx,6.5*\yy) {\sage{orange}{27}};
\node (v28) at (3*\xx,7.5*\yy) {\sage{magenta}{28}};
\node (v20) at (3*\xx,-0.5*\yy) {\sage{yellow}{20}};
\node (v19) at (3*\xx,-1.5*\yy) {\sage{green}{19}};
\node (v18) at (3*\xx,-2.5*\yy) {\sage{yellow}{18}};
\node (v17) at (3*\xx,-3.5*\yy) {\sage{green}{17}};
\node (v16) at (3*\xx,-4.5*\yy) {\sage{red}{16}};
\node (v15) at (3*\xx,-5.5*\yy) {\sage{cyan}{15}};
\node (v14) at (3*\xx,-6.5*\yy) {\sage{red}{14}};
\node (v13) at (3*\xx,-7.5*\yy) {\sage{cyan}{13}};
\node (v45) at (4*\xx,0.35*\yy) {\leaf};
\node (v46) at (4*\xx,0.65*\yy) {\leaf};
\node (v47) at (4*\xx,1.35*\yy) {\leaf};
\node (v48) at (4*\xx,1.65*\yy) {\leaf};
\node (v49) at (4*\xx,2.35*\yy) {\leaf};
\node (v50) at (4*\xx,2.65*\yy) {\leaf};
\node (v51) at (4*\xx,3.35*\yy) {\leaf};
\node (v52) at (4*\xx,3.65*\yy) {\leaf};
\node (v53) at (4*\xx,4.35*\yy) {\leaf};
\node (v54) at (4*\xx,4.65*\yy) {\leaf};
\node (v55) at (4*\xx,5.35*\yy) {\leaf};
\node (v56) at (4*\xx,5.65*\yy) {\leaf};
\node (v57) at (4*\xx,6.35*\yy) {\leaf};
\node (v58) at (4*\xx,6.65*\yy) {\leaf};
\node (v59) at (4*\xx,7.35*\yy) {\leaf};
\node (v60) at (4*\xx,7.65*\yy) {\leaf};
\node (v44) at (4*\xx,-0.35*\yy) {\leaf};
\node (v43) at (4*\xx,-0.65*\yy) {\leaf};
\node (v42) at (4*\xx,-1.35*\yy) {\leaf};
\node (v41) at (4*\xx,-1.65*\yy) {\leaf};
\node (v40) at (4*\xx,-2.35*\yy) {\leaf};
\node (v39) at (4*\xx,-2.65*\yy) {\leaf};
\node (v38) at (4*\xx,-3.35*\yy) {\leaf};
\node (v37) at (4*\xx,-3.65*\yy) {\leaf};
\node (v36) at (4*\xx,-4.35*\yy) {\leaf};
\node (v35) at (4*\xx,-4.65*\yy) {\leaf};
\node (v34) at (4*\xx,-5.35*\yy) {\leaf};
\node (v33) at (4*\xx,-5.65*\yy) {\leaf};
\node (v32) at (4*\xx,-6.35*\yy) {\leaf};
\node (v31) at (4*\xx,-6.65*\yy) {\leaf};
\node (v30) at (4*\xx,-7.35*\yy) {\leaf};
\node (v29) at (4*\xx,-7.65*\yy) {\leaf};
\draw[->] (v0) -- node [below, sloped] {\tiny{1st}} (v1);
\draw[->] (v0) -- node [below, sloped] {\tiny{2nd}} (v2);
\draw[->] (v0) --  node [above, sloped] {\tiny{3rd}} (v3);
\draw[->] (v0) --  node [above, sloped] {\tiny{Crew}} (v4);
\draw[->] (v1) --  node [below, sloped] {\tiny{Male}} (v5);
\draw[->] (v1) --  node [above, sloped] {\tiny{Female}} (v6);
\draw[->] (v2) --  node [below, sloped] {\tiny{Male}} (v7);
\draw[->] (v2) --  node [above, sloped] {\tiny{Female}} (v8);
\draw[->] (v3) --  node [below, sloped] {\tiny{Male}} (v9);
\draw[->] (v3) --  node [above, sloped] {\tiny{Female}} (v10);
\draw[->] (v4) --  node [below, sloped] {\tiny{Male}} (v11);
\draw[->] (v4) --  node [above, sloped] {\tiny{Female}} (v12);
\draw[->] (v5) --  node [below, sloped] {\tiny{No}} (v13);
\draw[->] (v5) --  node [above, sloped] {\tiny{Yes}} (v14);
\draw[->] (v6) --  node [below, sloped] {\tiny{No}} (v15);
\draw[->] (v6) --  node [above, sloped] {\tiny{Yes}} (v16);
\draw[->] (v7) --  node [below, sloped] {\tiny{No}} (v17);
\draw[->] (v7) --  node [above, sloped] {\tiny{Yes}} (v18);
\draw[->] (v8) --  node [below, sloped] {\tiny{No}} (v19);
\draw[->] (v8) --  node [above, sloped] {\tiny{Yes}} (v20);
\draw[->] (v9) --  node [below, sloped] {\tiny{No}} (v21);
\draw[->] (v9) --  node [above, sloped] {\tiny{Yes}} (v22);
\draw[->] (v10) --  node [below, sloped] {\tiny{No}} (v23);
\draw[->] (v10) --  node [above, sloped] {\tiny{Yes}} (v24);
\draw[->] (v11) --  node [below, sloped] {\tiny{No}} (v25);
\draw[->] (v11) --  node [above, sloped] {\tiny{Yes}} (v26);
\draw[->] (v12) --  node [below, sloped] {\tiny{No}} (v27);
\draw[->] (v12) --  node [above, sloped] {\tiny{Yes}} (v28);
\draw[->] (v13) --  node [below, sloped] {\tiny{Child}} (v29);
\draw[->] (v13) --  node [above, sloped] {\tiny{Adult}} (v30);
\draw[->] (v14) --  node [below, sloped] {\tiny{Child}} (v31);
\draw[->] (v14) --  node [above, sloped] {\tiny{Adult}} (v32);
\draw[->] (v15) --  node [below, sloped] {\tiny{Child}} (v33);
\draw[->] (v15) --  node [above, sloped] {\tiny{Adult}} (v34);
\draw[->] (v16) --  node [below, sloped] {\tiny{Child}} (v35);
\draw[->] (v16) --  node [above, sloped] {\tiny{Adult}} (v36);
\draw[->] (v17) --  node [below, sloped] {\tiny{Child}} (v37);
\draw[->] (v17) --  node [above, sloped] {\tiny{Adult}} (v38);
\draw[->] (v18) --  node [below, sloped] {\tiny{Child}} (v39);
\draw[->] (v18) --  node [above, sloped] {\tiny{Adult}} (v40);
\draw[->] (v19) --  node [below, sloped] {\tiny{Child}} (v41);
\draw[->] (v19) --  node [above, sloped] {\tiny{Adult}} (v42);
\draw[->] (v20) --  node [below, sloped] {\tiny{Child}} (v43);
\draw[->] (v20) --  node [above, sloped] {\tiny{Adult}} (v44);
\draw[->] (v21) --  node [below, sloped] {\tiny{Child}} (v45);
\draw[->] (v21) --  node [above, sloped] {\tiny{Adult}} (v46);
\draw[->] (v22) --  node [below, sloped] {\tiny{Child}} (v47);
\draw[->] (v22) --  node [above, sloped] {\tiny{Adult}} (v48);
\draw[->] (v23) --  node [below, sloped] {\tiny{Child}} (v49);
\draw[->] (v23) --  node [above, sloped] {\tiny{Adult}} (v50);
\draw[->] (v24) --  node [below, sloped] {\tiny{Child}} (v51);
\draw[->] (v24) --  node [above, sloped] {\tiny{Adult}} (v52);
\draw[->] (v25) --  node [below, sloped] {\tiny{Child}} (v53);
\draw[->] (v25) --  node [above, sloped] {\tiny{Adult}} (v54);
\draw[->] (v26) --  node [below, sloped] {\tiny{Child}} (v55);
\draw[->] (v26) --  node [above, sloped] {\tiny{Adult}} (v56);
\draw[->] (v27) --  node [below, sloped] {\tiny{Child}} (v57);
\draw[->] (v27) --  node [above, sloped] {\tiny{Adult}} (v58);
\draw[->] (v28) --  node [below, sloped] {\tiny{Child}} (v59);
\draw[->] (v28) --  node [above, sloped] {\tiny{Adult}} (v60);
\end{tikzpicture}
}

\caption{The staged tree representation of the BN in Figure \ref{fig:bn} for the \texttt{Titanic} dataset. \label{fig:staged2}}
\end{figure}
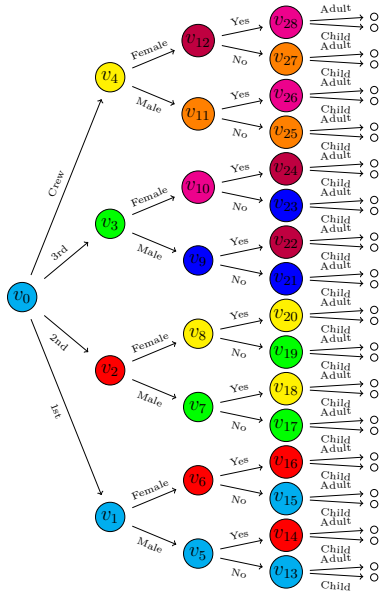

As an illustration, Figure \ref{fig:staged2} reports the tree $T_G$ associated to the BN in Figure \ref{fig:bn}.  Since the variables $\texttt{Class}$, $\texttt{Gender}$ and $\texttt{Survived}$ are fully connected in the BN, the associated staged tree is such that vertices at depth one and two are in their own individual stages. The only symmetric conditional independence embedded in the BN is represented by joining pairs of vertices at depth three (associated to the variable \texttt{Age)} in the same stage. Clearly, the staging of the staged tree representing a BN in Figure \ref{fig:staged2} exhibits a lot more symmetry than the one in Figure \ref{fig:staged1}, that can represent a wide array of non-symmetric independences.

Our first contribution is the solution of the following inverse problem: given an $\bf X$-compatible  staged tree $T=(V, E, \theta)$ find the corresponding DAG $G$.  
This DAG cannot represent, in general, 
the same staged tree model, 
since BNs cannot represent 
non-symmetric conditional independences. 
Nevertheless, we prove that we can retrieve a 
minimal DAG, in a sense the we formalize next. 
A proof of the result is in the supplementary material. 

\begin{proposition}
\label{theo:1}
	Let $T = (V, E, \theta)$ be an $\bf X$-compatible
	staged tree, with $\kappa: V \to \mathcal{C}$ the vertex labeling that 
	defines $\theta$. 
	Let 
	$G_T = ([p], F_T)$ be the DAG with vertex set $[p]$  and whose edge 
	set $F_T$ includes 
	the edge $(k,i), k < i$, if and only if 
	there exist $\mathbf{x}_{[i-1]}, \mathbf{x'}_{[i-1]} \in \mathbb{X}_{[i-1]}$ 
		such that $x_j = x'_j$ for all $j \neq k$ and 
\begin{equation}
\label{eq:theo}
	\kappa(\mathbf{x}_{[i-1]}) 
	 \neq \kappa(\mathbf{x'}_{[i-1]}).
\end{equation}
Then $G_T = ([p], F_T)$ is the minimal DAG such that 
	 $\mathcal{M}_T \subseteq \mathcal{M}_{G_T}$, in the sense that 
	 for every DAG $G = ([p],F)$ such that $1, \ldots, p$ is 
	 a topological order, 
	 if $\mathcal{M}_T \subseteq \mathcal{M}_{G}$ then $F_T \subseteq F$. 	In particular $X_A \independent X_B | X_C$ holds in 
	$\mathcal{M}_T$ if and only if 
	$A$ and $B$ are  d-separated by $C$ in $G_T$. 
\end{proposition}

\begin{corollary}
\label{cor}
In the setup of Proposition \ref{theo:1}, $
\mathcal{M}_{T_G}=\mathcal{M}_{G_{T_G}}$.
\end{corollary}

A staged tree $T$ is therefore a sub-model of the resulting $G_T$ which embeds the same set of symmetric conditional independences. The BN $G_T$ is minimal in the sense that it includes the smallest number of edges among all possible BNs that include $\mathcal{M}_T$ as a sub-model. The models  $\mathcal{M}_T$ and $\mathcal{M}_{G_T}$ are equal if and only if $T$ embeds only symmetric conditional independences. As an illustration consider the staged tree in Figure \ref{fig:staged1}. It can be shown using Proposition \ref{theo:1} that the associated BN $G_T$ is complete and therefore it must be that $\mathcal{M}_T\subseteq \mathcal{M}_{G_T}$. Conversely, if the staged tree in Figure \ref{fig:staged2} is transformed into a BN, then using Corollary \ref{cor} the resulting BN must be the one in Figure \ref{fig:bn}.

Importantly, Theorem \ref{theo:1} gives a novel criterion to read symmetric conditional independence statements from a staged tree, by transforming it into a BN whose structure represents the same equalities of the form in (\ref{ci2}). Conditional independence statements in the staged tree can then be read from the associated BN using the d-separation criterion \citep[see e.g.][]{Darwiche2009}. For instance, the staged tree in Figure \ref{fig:staged1} does not embed any symmetric conditional independence, since the associated BN is complete.


The supplementary material gives a detailed implementation of both conversion algorithms, from BN to staged tree and vice versa.

\section{Non-Symmetric Dependence and DAGs}

Proposition \ref{theo:1} identifies if there is a dependence between two random variables in a $\bf X$-compatible staged tree $T$ and in such a case draws an edge in $G_T$. However, the staged tree carries a lot more information about the type of relationship existing between the two variables. In this section we introduce methods to label the edges of $G_T$ so to depict some of the information about the non-symmetric independences of $T$ in $G_T$.

\subsection{Classes of Statistical Dependence}
 
First we need to characterize the type of dependence existing between 
two random variables that are joined by an edge in a DAG $G$.

\begin{definition}
\label{def:class}
		Let  $P$ be  the joint
	mass function of $\mathbf{X}$
	and  $P$ be Markov with respect to a DAG $G= ([p], F)$. 
		For each $(j,i) \in F$
	        we say that the dependence of $X_i$ from $X_j$ is of class
\begin{itemize}
\item \emph{context}, if $X_i$ and $X_j$ are context-specific independent given 
	some context $\mathbf{x}_{C}$ with $C =  \Pi_i \setminus \{j\}$. 
\item \emph{partial}, if $X_i$ is partially conditionally independent of $X_j$ in a
	domain $\mathcal{D}_j \subset \mathbb{X}_j$ given 
	a context $\mathbf{x}_{C}$ with $C =  \Pi_i \setminus \{j\}$; and
		$X_i$ and $X_j$ are not context-specific independent given the same context 
		$\mathbf{x}_C$. 
	\item \emph{local}, if none of the above hold and a local independence of the 
		form $P(x_i| \mathbf{x}_{\Pi_i}) = P(x_i| \tilde{\mathbf{x}}_{\Pi_i})$ is valid 
		where $x_j \neq \tilde{x}_j$.
\item \emph{total}, if none of the above hold. 
\end{itemize}



\end{definition}

Notice that if the class of dependence between $X_i$ and $X_j$ is context or
partial then there may also be  local independence statements as in
(\ref{eq:lci}) involving these two variables. 
Similarly, the dependence between $X_i$ and $X_j$ can be both context and
partial with respect to two different contexts. 
On the other hand if their class
of dependence is local then, by definition,
there are no context-specific or partial equalities.

Proposition \ref{theo:1} paves the way to assess the class of dependence
existing between $X_i$ and $X_j$. In particular, one has to check if there are
equalities of the form (\ref{eq:theo})  for all or some
$\bm{x}_{[i]}\in\mathbb{X}_{[i]}$ and, if so, to which class they correspond. A
discussion of the implementation of such checks is in the supplementary material. As illustrated in Section \ref{sec:5}, these can be performed quickly although all combinations of ancestral variables have to be considered.

\subsection{Asymmetry-Labeled DAGs
}
\label{sec:albn}
An edge in a  BN represents, by construction, a total dependence between
two random variables.  However, the flexibility of staged trees allows us to
assess if such a dependence is of any other of the classes introduced in
Definition \ref{def:class}. This observation leads us to define a 
new graphical representation, 
that we term 
\emph{asymmetry-labeled DAG} (ALDAG), where
edges are colored depending on the type of relationship between variables. 
 
Formally, let $G$ be a DAG and $F$ its edge set and
\[\mathcal{L}^A=\{\textnormal{`context'}, \textnormal{`partial'}, \textnormal{`context/partial'}, \textnormal{`local'}, \textnormal{`total'} \}\] 
be the set of edge labels marking the type of dependence.

\begin{definition}
	An ALDAG is a pair $(G,\psi)$ where $G=([p], F)$ is a DAG and $\psi$ is a function from the edge set of $G$ to $\mathcal{L}^A$, i.e. $\psi: F\rightarrow \mathcal{L}^A$. We say that a joint mass function $P$ is compatible with an ALDAG 
	$(G, \psi)$ if $P$ is Markov to $G$ and additionally 
	$P$ respects all the edge labels given by $\psi$; that is,
	for each $(j,i) \in F$, $X_i$ is 
	$\psi(i,j)$ dependent from $X_j$.   
\end{definition}

Henceforth, we represent the labeling via a coloring of the edges of the ALDAG. Standard BNs have an ALDAG representation where all edges have label `total'. Notice that standard features of BNs are also valid over ALDAGs: for instance, the already-mentioned d-separation criterion as well as fast probability propagation algorithms.

ALDAGs share features with labeled DAGs of \cite{Pensar2015} but they differ in two critical aspects: first, labeled DAGs can only embed context-specific independence whilst ALDAGs represent any type of asymmetric independence; second, labeled DAGs specifically report the contexts over which independences hold, whilst ALDAGs do not. There are two reasons behind this: on one hand, the specific independences in ALDAGs can be read from the associated staged tree; on the other, for applications with a larger number of variables the required contexts are often too complex to be reported within the DAG.

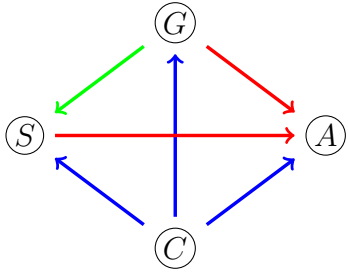
\begin{figure}
\centering
\begin{tikzpicture}
\renewcommand{\xx}{2}
\renewcommand{\yy}{1.5}
\node (1) at (1*\xx,-1*\yy){\stages{white}{C}};
\node (2) at (1*\xx,1*\yy){\stages{white}{G}};
\node (3) at (0*\xx,0*\yy){\stages{white}{S}};
\node (4) at (2*\xx,0*\yy){\stages{white}{A}};
\draw[->,blue, line width = 1.3pt] (1) -- (2);
\draw[->,blue, line width = 1.3pt] (1) -- (3);
\draw[->,blue, line width = 1.3pt] (1) -- (4);
\draw[->,green, line width = 1.3pt] (2) -- (3);
\draw[->,red, line width = 1.3pt] (2) -- (4);
\draw[->,red, line width = 1.3pt] (3) -- (4);
\end{tikzpicture}
\caption{An ALDAG for the \texttt{Titanic} dataset constructed from the staged tree in Figure \ref{fig:staged1}. The edge coloring is: red - context; blue - partial; green - local.\label{fig:albn}}
\end{figure}

As an illustration of ALDAGs  consider the staged tree for the \texttt{Titanic} data in Figure \ref{fig:staged1} which, using Proposition \ref{theo:1}, is transformed into the ALDAG in Figure \ref{fig:albn}. Although the ALDAG does not carry all the information stored in the staged tree, which is quite complex to read, it intuitively describes the classes of dependence existing among the random variables. The blue edges denote that there is partial dependence between \texttt{Class} and any other variable. The red edges denote that \texttt{Age} has a context dependence with \texttt{Gender} and \texttt{Survived}. Notice importantly that in the standard BN, which does not have the flexibility to embed non-symmetric independences, the variables \texttt{Age} and \texttt{Gender} were considered conditionally independent. Lastly, the green edge between \texttt{Gender} and \texttt{Survived} implies that there is only a local dependence between these two variables, and therefore there is no specific pattern guiding the equalities of probabilities between these two variables. Such extended forms of dependence better describe the fate of the Titanic passengers since, as already noticed, the BIC of the associated staged tree is smaller than the one of the best scoring BN.

\subsection{Constructing ALDAGs}
\label{sec:learning}

ALDAGs can be obtained from estimated staged trees,
and, in particular, with the following routine:
(i) learn a staged tree model $T$ from data, 
using for instance any of the algorithms in 
\texttt{stagedtrees}; 
(ii) transform $T$ into $G_T$ as in Proposition \ref{theo:1}; 
(iii) assign a label to each edge of $G_T$ by checking the equalities in (\ref{eq:theo}) that hold in $T$. 
Steps (ii) and (iii) are implemented in the \texttt{stagedtrees} R package using the algorithms in the supplementary material.

The most critical and computationally expensive step of learning an ALDAG is the staged tree learning step (i). 
There is a now large literature on learning staged trees from data \citep{Carli2020a,Carli2020,Cowell2014,Freeman2011,leonelli2022structural,Silander2013}. 
Here we consider the available algorithms 
implemented in the 
\texttt{stagedtrees} R package~\citep{Carli2020}. 
In particular, we will use the following algorithms which work with a fixed order of the variables 
\citep[see][for more details]{Carli2020}:
\begin{itemize}
\item an hill-climbing (HC) algorithm which at each step either joins or splits vertices of the tree in stages by optimizing a model score (usually the BIC);
\item a backward (hill-climbing) (BHC) algorithm
which, at each step, can only join stages 
together by optimizing a model score. 
\item a novel backward algorithm which iteratively 
add context-specific independences (CSBHC, see Sec. \ref{sec:csbhc}). 
\end{itemize}
Furthermore, any of the above mentioned algorithms can be used within the dynamic programming approach of \citet{Cowell2014} to also choose an optimal order of the variables \citep{leonelli2021context}. 

An ALDAG can also be 
obtained
as a refinement of the DAG of a BN by the 
addition of edge labels indicating the class of dependence. 
Given a DAG $G$, the following steps implement such a refinement: 
(i) transform $G$ into the staged tree $T_G$ using any of the topological orders of variables; 
(ii) run a backward hill-climbing algorithm  using $T_G$ as starting model and obtain a new tree $T$; 
(iii) transform $T$ into $G_{T}$ and apply the edge-labeling. 
The resulting ALDAG has an edge set which is either equal to or a subset of the edge set of $G$. Furthermore, the edge set is now labeled and denoting the classes of dependence. 

The use of these algorithms in practice is extensively illustrated in Section \ref{sec:5} below.

\subsubsection{Searching Context-Specific Independences}
\label{sec:csbhc}
The flexibility of staged trees to represent any type of non-symmetric independence has two major drawbacks: on one hand reading independences from the tree can become complex; on the other learning trees from data can be computationally very expensive. 
To address these two difficulties we introduce here a
new heuristic search for the stage structure, motivated by our new definition of the ALDAG.
In particular we consider a backward hill-climbing 
algorithms that, for each variable, 
iteratively adds context-specific independence 
relationships to optimize a given score (e.g. BIC). 
In particular, at each step, the algorithm 
searches all possible additional 
context-specific conditional independences 
of the form, 
\[  X_i \independent X_j |X_C= x_{C} \quad j < i  \]
where $C = [i] \setminus \{i,j\} $ and thus 
$x_{C} \in \mathbb{X}_C$ is a context specified 
by all variables preceding $X_i$ in the tree except $X_j$. Complete details about the algorithm can be found in the supplementary material.


\begin{figure}
\centering
\scalebox{0.65}{
\begin{tikzpicture}
\renewcommand{\xx}{1.8}
\renewcommand{\yy}{0.75}
\node (v0) at (0*\xx,0*\yy) {\sage{cyan}{0}};
\node (v1) at (1*\xx,-6*\yy) {\sage{cyan}{1}};
\node (v2) at (1*\xx,-2*\yy) {\sage{cyan}{2}};
\node (v3) at (1*\xx,2*\yy) {\sage{green}{3}};
\node (v4) at (1*\xx,6*\yy) {\sage{yellow}{4}};
\node (v9) at (2*\xx,1*\yy) {\sage{blue}{9}};
\node (v10) at (2*\xx,3*\yy) {\sage{magenta}{10}};
\node (v11) at (2*\xx,5*\yy) {\sage{orange}{11}};
\node (v12) at (2*\xx,7*\yy) {\sage{purple}{12}};
\node (v8) at (2*\xx,-1*\yy) {\sage{yellow}{8}};
\node (v7) at (2*\xx,-3*\yy) {\sage{green}{7}};
\node (v6) at (2*\xx,-5*\yy) {\sage{red}{6}};
\node (v5) at (2*\xx,-7*\yy) {\sage{cyan}{5}};
\node (v21) at (3*\xx,0.5*\yy) {\sage{red}{21}};
\node (v22) at (3*\xx,1.5*\yy) {\sage{red}{22}};
\node (v23) at (3*\xx,2.5*\yy) {\sage{red}{23}};
\node (v24) at (3*\xx,3.5*\yy) {\sage{red}{24}};
\node (v25) at (3*\xx,4.5*\yy) {\sage{magenta}{25}};
\node (v26) at (3*\xx,5.5*\yy) {\sage{magenta}{26}};
\node (v27) at (3*\xx,6.5*\yy) {\sage{red}{27}};
\node (v28) at (3*\xx,7.5*\yy) {\sage{magenta}{28}};
\node (v20) at (3*\xx,-0.5*\yy) {\sage{red}{20}};
\node (v19) at (3*\xx,-1.5*\yy) {\sage{red}{19}};
\node (v18) at (3*\xx,-2.5*\yy) {\sage{yellow}{18}};
\node (v17) at (3*\xx,-3.5*\yy) {\sage{green}{17}};
\node (v16) at (3*\xx,-4.5*\yy) {\sage{cyan}{16}};
\node (v15) at (3*\xx,-5.5*\yy) {\sage{red}{15}};
\node (v14) at (3*\xx,-6.5*\yy) {\sage{cyan}{14}};
\node (v13) at (3*\xx,-7.5*\yy) {\sage{cyan}{13}};
\node (v45) at (4*\xx,0.35*\yy) {\leaf};
\node (v46) at (4*\xx,0.65*\yy) {\leaf};
\node (v47) at (4*\xx,1.35*\yy) {\leaf};
\node (v48) at (4*\xx,1.65*\yy) {\leaf};
\node (v49) at (4*\xx,2.35*\yy) {\leaf};
\node (v50) at (4*\xx,2.65*\yy) {\leaf};
\node (v51) at (4*\xx,3.35*\yy) {\leaf};
\node (v52) at (4*\xx,3.65*\yy) {\leaf};
\node (v53) at (4*\xx,4.35*\yy) {\leaf};
\node (v54) at (4*\xx,4.65*\yy) {\leaf};
\node (v55) at (4*\xx,5.35*\yy) {\leaf};
\node (v56) at (4*\xx,5.65*\yy) {\leaf};
\node (v57) at (4*\xx,6.35*\yy) {\leaf};
\node (v58) at (4*\xx,6.65*\yy) {\leaf};
\node (v59) at (4*\xx,7.35*\yy) {\leaf};
\node (v60) at (4*\xx,7.65*\yy) {\leaf};
\node (v44) at (4*\xx,-0.35*\yy) {\leaf};
\node (v43) at (4*\xx,-0.65*\yy) {\leaf};
\node (v42) at (4*\xx,-1.35*\yy) {\leaf};
\node (v41) at (4*\xx,-1.65*\yy) {\leaf};
\node (v40) at (4*\xx,-2.35*\yy) {\leaf};
\node (v39) at (4*\xx,-2.65*\yy) {\leaf};
\node (v38) at (4*\xx,-3.35*\yy) {\leaf};
\node (v37) at (4*\xx,-3.65*\yy) {\leaf};
\node (v36) at (4*\xx,-4.35*\yy) {\leaf};
\node (v35) at (4*\xx,-4.65*\yy) {\leaf};
\node (v34) at (4*\xx,-5.35*\yy) {\leaf};
\node (v33) at (4*\xx,-5.65*\yy) {\leaf};
\node (v32) at (4*\xx,-6.35*\yy) {\leaf};
\node (v31) at (4*\xx,-6.65*\yy) {\leaf};
\node (v30) at (4*\xx,-7.35*\yy) {\leaf};
\node (v29) at (4*\xx,-7.65*\yy) {\leaf};
\draw[->] (v0) -- node [below, sloped] {\tiny{1st}} (v1);
\draw[->] (v0) -- node [below, sloped] {\tiny{2nd}} (v2);
\draw[->] (v0) --  node [above, sloped] {\tiny{3rd}} (v3);
\draw[->] (v0) --  node [above, sloped] {\tiny{Crew}} (v4);
\draw[->] (v1) --  node [below, sloped] {\tiny{Male}} (v5);
\draw[->] (v1) --  node [above, sloped] {\tiny{Female}} (v6);
\draw[->] (v2) --  node [below, sloped] {\tiny{Male}} (v7);
\draw[->] (v2) --  node [above, sloped] {\tiny{Female}} (v8);
\draw[->] (v3) --  node [below, sloped] {\tiny{Male}} (v9);
\draw[->] (v3) --  node [above, sloped] {\tiny{Female}} (v10);
\draw[->] (v4) --  node [below, sloped] {\tiny{Male}} (v11);
\draw[->] (v4) --  node [above, sloped] {\tiny{Female}} (v12);
\draw[->] (v5) --  node [below, sloped] {\tiny{No}} (v13);
\draw[->] (v5) --  node [above, sloped] {\tiny{Yes}} (v14);
\draw[->] (v6) --  node [below, sloped] {\tiny{No}} (v15);
\draw[->] (v6) --  node [above, sloped] {\tiny{Yes}} (v16);
\draw[->] (v7) --  node [below, sloped] {\tiny{No}} (v17);
\draw[->] (v7) --  node [above, sloped] {\tiny{Yes}} (v18);
\draw[->] (v8) --  node [below, sloped] {\tiny{No}} (v19);
\draw[->] (v8) --  node [above, sloped] {\tiny{Yes}} (v20);
\draw[->] (v9) --  node [below, sloped] {\tiny{No}} (v21);
\draw[->] (v9) --  node [above, sloped] {\tiny{Yes}} (v22);
\draw[->] (v10) --  node [below, sloped] {\tiny{No}} (v23);
\draw[->] (v10) --  node [above, sloped] {\tiny{Yes}} (v24);
\draw[->] (v11) --  node [below, sloped] {\tiny{No}} (v25);
\draw[->] (v11) --  node [above, sloped] {\tiny{Yes}} (v26);
\draw[->] (v12) --  node [below, sloped] {\tiny{No}} (v27);
\draw[->] (v12) --  node [above, sloped] {\tiny{Yes}} (v28);
\draw[->] (v13) --  node [below, sloped] {\tiny{Child}} (v29);
\draw[->] (v13) --  node [above, sloped] {\tiny{Adult}} (v30);
\draw[->] (v14) --  node [below, sloped] {\tiny{Child}} (v31);
\draw[->] (v14) --  node [above, sloped] {\tiny{Adult}} (v32);
\draw[->] (v15) --  node [below, sloped] {\tiny{Child}} (v33);
\draw[->] (v15) --  node [above, sloped] {\tiny{Adult}} (v34);
\draw[->] (v16) --  node [below, sloped] {\tiny{Child}} (v35);
\draw[->] (v16) --  node [above, sloped] {\tiny{Adult}} (v36);
\draw[->] (v17) --  node [below, sloped] {\tiny{Child}} (v37);
\draw[->] (v17) --  node [above, sloped] {\tiny{Adult}} (v38);
\draw[->] (v18) --  node [below, sloped] {\tiny{Child}} (v39);
\draw[->] (v18) --  node [above, sloped] {\tiny{Adult}} (v40);
\draw[->] (v19) --  node [below, sloped] {\tiny{Child}} (v41);
\draw[->] (v19) --  node [above, sloped] {\tiny{Adult}} (v42);
\draw[->] (v20) --  node [below, sloped] {\tiny{Child}} (v43);
\draw[->] (v20) --  node [above, sloped] {\tiny{Adult}} (v44);
\draw[->] (v21) --  node [below, sloped] {\tiny{Child}} (v45);
\draw[->] (v21) --  node [above, sloped] {\tiny{Adult}} (v46);
\draw[->] (v22) --  node [below, sloped] {\tiny{Child}} (v47);
\draw[->] (v22) --  node [above, sloped] {\tiny{Adult}} (v48);
\draw[->] (v23) --  node [below, sloped] {\tiny{Child}} (v49);
\draw[->] (v23) --  node [above, sloped] {\tiny{Adult}} (v50);
\draw[->] (v24) --  node [below, sloped] {\tiny{Child}} (v51);
\draw[->] (v24) --  node [above, sloped] {\tiny{Adult}} (v52);
\draw[->] (v25) --  node [below, sloped] {\tiny{Child}} (v53);
\draw[->] (v25) --  node [above, sloped] {\tiny{Adult}} (v54);
\draw[->] (v26) --  node [below, sloped] {\tiny{Child}} (v55);
\draw[->] (v26) --  node [above, sloped] {\tiny{Adult}} (v56);
\draw[->] (v27) --  node [below, sloped] {\tiny{Child}} (v57);
\draw[->] (v27) --  node [above, sloped] {\tiny{Adult}} (v58);
\draw[->] (v28) --  node [below, sloped] {\tiny{Child}} (v59);
\draw[->] (v28) --  node [above, sloped] {\tiny{Adult}} (v60);
\end{tikzpicture}
}
\;\;\;\;\;\;\;\;\;\;\;\;\;\;\;\;
\begin{tikzpicture}
\renewcommand{\xx}{2}
\renewcommand{\yy}{1.5}
\node (1) at (1*\xx,-1*\yy){\stages{white}{C}};
\node (2) at (1*\xx,1*\yy){\stages{white}{G}};
\node (3) at (0*\xx,0*\yy){\stages{white}{S}};
\node (4) at (2*\xx,0*\yy){\stages{white}{A}};
\draw[->,blue, line width = 1.3pt] (1) -- (2);
\draw[->,black, line width = 1.3pt] (1) -- (3);
\draw[->,violet, line width = 1.3pt] (1) -- (4);
\draw[->,black, line width = 1.3pt] (2) -- (3);
\draw[->,red, line width = 1.3pt] (2) -- (4);
\draw[->,red, line width = 1.3pt] (3) -- (4);
\end{tikzpicture}

\caption{A 
staged tree
learned with the proposed CSBHC algorithm over the \texttt{Titanic} dataset and its associated ALDAG. The edge coloring is: red - context; blue - partial; purple - context/partial; black - total.\label{fig:pitree}}
\end{figure}

Therefore the ALDAG of a staged tree learned 
with the CSBHC algorithm,
can only have context and partial (and context/partial) edge labels. 
As an example, in Figure \ref{fig:pitree} we report
the tree learned with CSBHC for the Titanic dataset
and its associated ALDAG. 
It can be seen that the ALDAG shares some features
with the one in Figure \ref{fig:albn}, 
but in this case the local edge does not appear. 
The tree in Figure~\ref{fig:pitree} has a BIC of 10479 which is worse than the one of the generic staged tree (BIC = 10440), 
but still better than the BN (BIC = 10502), again highlighting the need for models embedding non-symmetric independences.

\section{Applications}
\label{sec:5}

We now consider a variety of datasets commonly used in the probabilistic graphical models literature. First we carry out an experiment to assess the performance of ALDAGs as well as the complexity of our routines. Then we consider two additional real-world applications to further illustrate the capabilities of staged trees and ALDAGs.

\subsection{Computational Experiment}

Nine datasets which are either available in R packages or downloaded from the UCI repository are considered. For each dataset a DAG is first learned by
optimizing the BIC score \citep[using a tabu greedy search, ][]{Scutari2010} and then both the BHC and the proposed CSBHC algorithms are used to refine the DAG to staged trees. The learned staged trees are then transformed into ALDAGs using Algorithm~\ref{alg:staged_to_bn} given in the supplementary material. The results, summarized in Tables \ref{tab:1} and \ref{tab:times}, suggest the following:

\begin{table}

\caption{Results of the data experiments. ALDAG-$*$ columns report the number of edges by type (total, context, partial, context/partial, local)
in the two ALDAGs obtained by refinement of the DAG with BHC and CSBHC algorithms respectively. 
}
\label{tab:1}
\centering

\scalebox{0.75}{
\begin{tabular}{lrllrrr}
  \toprule
 & variables & ALDAG-BHC & ALDAG-CSBHC & BIC DAG & BIC BHC & BIC CSBHC \\ 
  \midrule
abalone & 9 & (5,10,0,0,6) & (10,11,0,0,0) & 29090 & 28895 & 28925 \\ 
  breastcancer & 10 & (4,0,3,0,0) & (7,0,0,0,0) & 4644 & 4594 & 4644 \\ 
  creditapproval & 14 & (10,3,0,0,2) & (10,5,0,0,0) & 9881 & 9862 & 9863 \\ 
  housevotes & 17 & (9,11,0,0,6) & (11,15,0,0,0) & 3539 & 3481 & 3488 \\ 
  indianliver & 11 & (10,1,0,0,4) & (14,1,0,0,0) & 6984 & 6967 & 6979 \\ 
  nursery & 9 & (0,0,5,3,0) & (5,1,0,2,0) & 252566 & 251922 & 252389 \\ 
  phdarticles & 6 & (3,2,1,0,0) & (4,2,0,0,0) & 8372 & 8357 & 8358 \\ 
  tic-tac-toe & 10 & (1,0,12,0,0) & (9,4,0,0,0) & 19277 & 18996 & 19251 \\ 
  titanic & 4 & (0,1,3,0,1) & (4,1,0,0,0) & 10502 & 10452 & 10488 \\ 
   \bottomrule
\end{tabular}
}

\end{table}

\begin{table}
    \centering
        \caption{Elapsed time (in seconds) for the data experiments. The DAG column 
        reports the seconds  
        to estimate the DAG; The columns BHC and 
        CSBHC report the seconds needed to refine
        the DAG to a staged tree, with the BHC 
        and the CSBHC algorithms respectively; 
        lastly the ALDAG column contains the time
        to build the ALDAG from the staged
        trees. 
        }
        \label{tab:times}
\scalebox{0.8}{
\begin{tabular}{lrrrr}
  \toprule
 & DAG & BHC & CSBHC & ALDAG \\ 
  \midrule
abalone.rds & 0.038 & 0.080 & 0.600 & 0.008 \\ 
  breastcancer.rds & 0.011 & 1.511 & 23.813 & 0.251 \\ 
  creditapproval.rds & 0.018 & 1.788 & 140.392 & 0.348 \\ 
  housevotes.rds & 0.020 & 3.073 & 411.043 & 0.672 \\ 
  indianliver.rds & 0.018 & 0.064 & 0.451 & 0.018 \\ 
  nursery.rds & 0.064 & 1.688 & 140.416 & 0.113 \\ 
  phdarticles.rds & 0.012 & 0.013 & 0.045 & 0.002 \\ 
  tic-tac-toe.rds & 0.016 & 0.639 & 10.415 & 0.076 \\ 
  titanic.rds & 0.007 & 0.027 & 0.017 & 0.001 \\ 
   \bottomrule
\end{tabular}
}

\end{table}

\begin{itemize}
\item The 
refined staged tree
provides a better fit than the standard BN in all datasets considered, thus highlighting the need to consider models which embed asymmetric conditional independences to untangle complex dependence structure;
\item only a small fraction of the edges learned via BN structural search algorithms are actually related to a symmetric dependence between variables. All ALDAGs have a much larger number of edges with a label which is not total;
\item the construction of the ALDAG does not impose a computational burden with respect to the staged tree model selection step, which is in all cases but one, the most computationally expensive task. Furthermore, the experiment shows that the methods are efficiently implemented even for a medium-large number of variables.
\end{itemize}

\subsection{Aspects of Everyday Life}
We next illustrate the use of staged trees to uncover dependence structures using data from the 2014 survey “Aspects on everyday life" collected by ISTAT (the Italian National Institute of Statistics) \citep{ISTAT}. The survey collects information from the Italian population on a variety of aspects of their daily lives. For the purpose of this analysis we consider five of the many questions asked in the survey: do you practice sports regularly? (S = yes/no); do you have friends you can count on? (F =
yes/no/unsure); do you believe in people? (B = yes/no); what's your gender (G =
male/female); what grade would you give to your life? (L = low/medium/high)\footnote{The original grade is numeric between zero and ten and it has been aggregated as follows: 0-5/low; 6-8/medium; 9-10/high.}. Instances with a missing answer were dropped, resulting in 38156 answers to the survey. Our aim is to analyse how various factors affect the life grade of the Italian population.

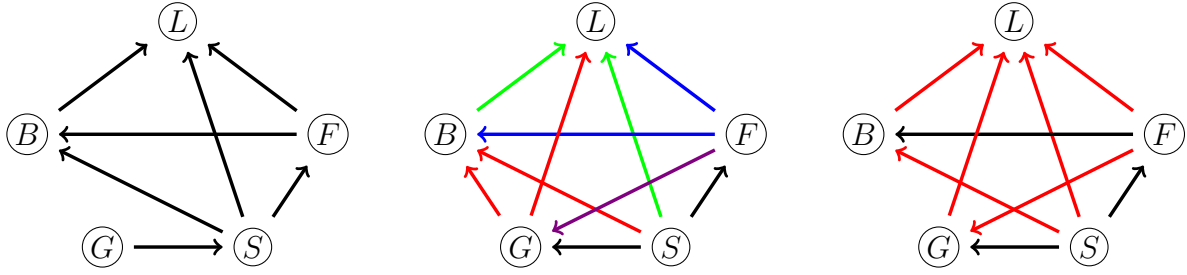
\begin{figure}
\begin{center}
\begin{tikzpicture}
\renewcommand{\xx}{1}
\renewcommand{\yy}{1.5}
\node (B) at (0*\xx,1*\yy){\stages{white}{B}};
\node (G) at (1*\xx,0*\yy){\stages{white}{G}};
\node (L) at (2*\xx,2*\yy){\stages{white}{L}};
\node (S) at (3*\xx,0*\yy){\stages{white}{S}};
\node (F) at (4*\xx,1*\yy){\stages{white}{F}};
\draw[->,black, line width = 1.3pt] (G) -- (S);
\draw[->,black, line width = 1.3pt] (S) -- (F);
\draw[->,black, line width = 1.3pt] (S) -- (L);
\draw[->,black, line width = 1.3pt] (S) -- (B);
\draw[->,black, line width = 1.3pt] (F) -- (B);
\draw[->,black, line width = 1.3pt] (F) -- (L);
\draw[->,black, line width = 1.3pt] (B) -- (L);
\end{tikzpicture}
\;\;\;\;
\begin{tikzpicture}
\renewcommand{\xx}{1}
\renewcommand{\yy}{1.5}
\node (B) at (0*\xx,1*\yy){\stages{white}{B}};
\node (G) at (1*\xx,0*\yy){\stages{white}{G}};
\node (L) at (2*\xx,2*\yy){\stages{white}{L}};
\node (S) at (3*\xx,0*\yy){\stages{white}{S}};
\node (F) at (4*\xx,1*\yy){\stages{white}{F}};
\draw[->,black, line width = 1.3pt] (S) -- (G);
\draw[->,black, line width = 1.3pt] (S) -- (F);
\draw[->,green, line width = 1.3pt] (S) -- (L);
\draw[->,red, line width = 1.3pt] (S) -- (B);
\draw[->,blue, line width = 1.3pt] (F) -- (B);
\draw[->,blue, line width = 1.3pt] (F) -- (L);
\draw[->,violet, line width = 1.3pt] (F) -- (G);
\draw[->,red, line width = 1.3pt] (G) -- (B);
\draw[->,red, line width = 1.3pt] (G) -- (L);
\draw[->,green, line width = 1.3pt] (B) -- (L);
\end{tikzpicture}
\;\;\;\;
\begin{tikzpicture}
\renewcommand{\xx}{1}
\renewcommand{\yy}{1.5}
\node (B) at (0*\xx,1*\yy){\stages{white}{B}};
\node (G) at (1*\xx,0*\yy){\stages{white}{G}};
\node (L) at (2*\xx,2*\yy){\stages{white}{L}};
\node (S) at (3*\xx,0*\yy){\stages{white}{S}};
\node (F) at (4*\xx,1*\yy){\stages{white}{F}};
\draw[->,black, line width = 1.3pt] (S) -- (G);
\draw[->,black, line width = 1.3pt] (S) -- (F);
\draw[->,red, line width = 1.3pt] (S) -- (L);
\draw[->,red, line width = 1.3pt] (S) -- (B);
\draw[->,black, line width = 1.3pt] (F) -- (B);
\draw[->,red, line width = 1.3pt] (F) -- (L);
\draw[->,red, line width = 1.3pt] (F) -- (G);
\draw[->,red, line width = 1.3pt] (G) -- (L);
\draw[->,red, line width = 1.3pt] (B) -- (L);
\end{tikzpicture}
\end{center}
\caption{BN for the aspects of everyday life data (left) as well as ALDAGs over the same data associated to the staged trees learned with BHC (center) and CSBHC  (right). The edge coloring is: red - context; blue - partial; violet - context/partial; green - local; black - total. \label{fig:life}}
\end{figure}

A BN with the variable life grade (L) as 
downstream variable 
is learned using the hill-climbing function (optimizing the BIC score) in the \texttt{bnlearn} package by 
blocklisting all outbond edges from L.
The learned DAG is reported in Figure \ref{fig:life} (left). 
This embeds the symmetric conditional independence L,B,F$\independent$G$|$S: given the level of sports activity, gender has no effect on life grade, on trust in people and on the availability of friends. The learned BN has a BIC of 251781.

\begin{figure}
\centering
\scalebox{0.65}{
\begin{tikzpicture}
\renewcommand{\xx}{2}
\renewcommand{\yy}{1}
\node (v0) at (0*\xx,0*\yy) {\sage{cyan}{0}};
\node (v1) at (1*\xx,-6*\yy) {\sage{cyan}{1}};
\node (v2) at (1*\xx,6*\yy) {\sage{red}{2}};
\node (v3) at (2*\xx,-10*\yy) {\sage{cyan}{3}};
\node (v4) at (2*\xx,-6*\yy) {\sage{cyan}{4}};
\node (v5) at (2*\xx,-2*\yy) {\sage{red}{5}};
\node (v6) at (2*\xx,2*\yy) {\sage{green}{6}};
\node (v7) at (2*\xx,6*\yy) {\sage{green}{7}};
\node (v8) at (2*\xx,10*\yy) {\sage{green}{8}};
\node (v9) at (3*\xx,-11*\yy) {\sage{cyan}{9}};
\node (v10) at (3*\xx,-9*\yy) {\sage{cyan}{10}};
\node (v11) at (3*\xx,-7*\yy) {\sage{red}{11}};
\node (v12) at (3*\xx,-5*\yy) {\sage{red}{12}};
\node (v13) at (3*\xx,-3*\yy) {\sage{green}{13}};
\node (v14) at (3*\xx,-1*\yy) {\sage{green}{14}};
\node (v15) at (3*\xx,1*\yy) {\sage{green}{15}};
\node (v16) at (3*\xx,3*\yy) {\sage{green}{16}};
\node (v17) at (3*\xx,5*\yy) {\sage{yellow}{17}};
\node (v18) at (3*\xx,7*\yy) {\sage{yellow}{18}};
\node (v19) at (3*\xx,9*\yy) {\sage{green}{19}};
\node (v20) at (3*\xx,11*\yy) {\sage{cyan}{20}};
\node (v21) at (4*\xx,-11.5*\yy) {\sage{cyan}{21}};
\node (v22) at (4*\xx,-10.5*\yy) {\sage{red}{22}};
\node (v23) at (4*\xx,-9.5*\yy) {\sage{cyan}{23}};
\node (v24) at (4*\xx,-8.5*\yy) {\sage{green}{24}};
\node (v25) at (4*\xx,-7.5*\yy) {\sage{green}{25}};
\node (v26) at (4*\xx,-6.5*\yy) {\sage{yellow}{26}};
\node (v27) at (4*\xx,-5.5*\yy) {\sage{green}{27}};
\node (v28) at (4*\xx,-4.5*\yy) {\sage{yellow}{28}};
\node (v29) at (4*\xx,-3.5*\yy) {\sage{cyan}{29}};
\node (v30) at (4*\xx,-2.5*\yy) {\sage{yellow}{30}};
\node (v31) at (4*\xx,-1.5*\yy) {\sage{cyan}{31}};
\node (v32) at (4*\xx,-0.5*\yy) {\sage{green}{32}};
\node (v33) at (4*\xx,0.5*\yy) {\sage{magenta}{33}};
\node (v34) at (4*\xx,1.5*\yy) {\sage{blue}{34}};
\node (v35) at (4*\xx,2.5*\yy) {\sage{red}{35}};
\node (v36) at (4*\xx,3.5*\yy) {\sage{blue}{36}};
\node (v37) at (4*\xx,4.5*\yy) {\sage{gray}{37}};
\node (v38) at (4*\xx,5.5*\yy) {\sage{orange}{38}};
\node (v39) at (4*\xx,6.5*\yy) {\sage{gray}{39}};
\node (v40) at (4*\xx,7.5*\yy) {\sage{orange}{40}};
\node (v41) at (4*\xx,8.5*\yy) {\sage{gray}{41}};
\node (v42) at (4*\xx,9.5*\yy) {\sage{orange}{42}};
\node (v43) at (4*\xx,10.5*\yy) {\sage{green}{43}};
\node (v44) at (4*\xx,11.5*\yy) {\sage{magenta}{44}};
\node (v45) at (5*\xx,-11.83*\yy) {\leaf};
\node (v46) at (5*\xx,-11.5*\yy) {\leaf};
\node (v47) at (5*\xx,-11.17*\yy) {\leaf};
\node (v48) at (5*\xx,-10.83*\yy) {\leaf};
\node (v49) at (5*\xx,-10.5*\yy) {\leaf};
\node (v50) at (5*\xx,-10.17*\yy) {\leaf};
\node (v51) at (5*\xx,-9.83*\yy) {\leaf};
\node (v52) at (5*\xx,-9.5*\yy) {\leaf};
\node (v53) at (5*\xx,-9.17*\yy) {\leaf};
\node (v54) at (5*\xx,-8.83*\yy) {\leaf};
\node (v55) at (5*\xx,-8.5*\yy) {\leaf};
\node (v56) at (5*\xx,-8.17*\yy) {\leaf};
\node (v57) at (5*\xx,-7.83*\yy) {\leaf};
\node (v58) at (5*\xx,-7.5*\yy) {\leaf};
\node (v59) at (5*\xx,-7.17*\yy) {\leaf};
\node (v61) at (5*\xx,-6.83*\yy) {\leaf};
\node (v62) at (5*\xx,-6.5*\yy) {\leaf};
\node (v63) at (5*\xx,-6.17*\yy) {\leaf};
\node (v64) at (5*\xx,-5.83*\yy) {\leaf};
\node (v65) at (5*\xx,-5.5*\yy) {\leaf};
\node (v66) at (5*\xx,-5.17*\yy) {\leaf};
\node (v67) at (5*\xx,-4.83*\yy) {\leaf};
\node (v68) at (5*\xx,-4.5*\yy) {\leaf};
\node (v69) at (5*\xx,-4.17*\yy) {\leaf};
\node (v70) at (5*\xx,-3.83*\yy) {\leaf};
\node (v71) at (5*\xx,-3.5*\yy) {\leaf};
\node (v72) at (5*\xx,-3.17*\yy) {\leaf};
\node (v73) at (5*\xx,-2.83*\yy) {\leaf};
\node (v74) at (5*\xx,-2.5*\yy) {\leaf};
\node (v75) at (5*\xx,-2.17*\yy) {\leaf};
\node (v76) at (5*\xx,-1.83*\yy) {\leaf};
\node (v77) at (5*\xx,-1.5*\yy) {\leaf};
\node (v78) at (5*\xx,-1.17*\yy) {\leaf};
\node (v79) at (5*\xx,-0.83*\yy) {\leaf};
\node (v80) at (5*\xx,-0.5*\yy) {\leaf};
\node (v81) at (5*\xx,-0.17*\yy) {\leaf};
\node (v117) at (5*\xx,11.83*\yy) {\leaf};
\node (v116) at (5*\xx,11.5*\yy) {\leaf};
\node (v115) at (5*\xx,11.17*\yy) {\leaf};
\node (v114) at (5*\xx,10.83*\yy) {\leaf};
\node (v113) at (5*\xx,10.5*\yy) {\leaf};
\node (v112) at (5*\xx,10.17*\yy) {\leaf};
\node (v111) at (5*\xx,9.83*\yy) {\leaf};
\node (v110) at (5*\xx,9.5*\yy) {\leaf};
\node (v109) at (5*\xx,9.17*\yy) {\leaf};
\node (v108) at (5*\xx,8.83*\yy) {\leaf};
\node (v107) at (5*\xx,8.5*\yy) {\leaf};
\node (v106) at (5*\xx,8.17*\yy) {\leaf};
\node (v105) at (5*\xx,7.83*\yy) {\leaf};
\node (v104) at (5*\xx,7.5*\yy) {\leaf};
\node (v103) at (5*\xx,7.17*\yy) {\leaf};
\node (v102) at (5*\xx,6.83*\yy) {\leaf};
\node (v101) at (5*\xx,6.5*\yy) {\leaf};
\node (v100) at (5*\xx,6.17*\yy) {\leaf};
\node (v99) at (5*\xx,5.83*\yy) {\leaf};
\node (v98) at (5*\xx,5.5*\yy) {\leaf};
\node (v97) at (5*\xx,5.17*\yy) {\leaf};
\node (v96) at (5*\xx,4.83*\yy) {\leaf};
\node (v95) at (5*\xx,4.5*\yy) {\leaf};
\node (v94) at (5*\xx,4.17*\yy) {\leaf};
\node (v93) at (5*\xx,3.83*\yy) {\leaf};
\node (v92) at (5*\xx,3.5*\yy) {\leaf};
\node (v91) at (5*\xx,3.17*\yy) {\leaf};
\node (v90) at (5*\xx,2.83*\yy) {\leaf};
\node (v89) at (5*\xx,2.5*\yy) {\leaf};
\node (v88) at (5*\xx,2.17*\yy) {\leaf};
\node (v87) at (5*\xx,1.83*\yy) {\leaf};
\node (v86) at (5*\xx,1.5*\yy) {\leaf};
\node (v85) at (5*\xx,1.17*\yy) {\leaf};
\node (v84) at (5*\xx,0.83*\yy) {\leaf};
\node (v83) at (5*\xx,0.5*\yy) {\leaf};
\node (v82) at (5*\xx,0.17*\yy) {\leaf};
\draw[->] (v0) -- node [below, sloped] {\footnotesize{No}} (v1);
\draw[->] (v0) -- node [above, sloped] {\footnotesize{Yes}} (v2);
\draw[->] (v1) --  node [below, sloped] {\footnotesize{No}} (v3);
\draw[->] (v1) --  node [above, sloped] {\footnotesize{Yes}} (v4);
\draw[->] (v1) --  node [above, sloped] {\footnotesize{Unsure}} (v5);
\draw[->] (v2) --  node [below, sloped] {\footnotesize{No}} (v6);
\draw[->] (v2) --  node [above, sloped] {\footnotesize{Yes}} (v7);
\draw[->] (v2) --  node [above, sloped] {\footnotesize{Unsure}} (v8);
\draw[->] (v3) --  node [below, sloped] {\footnotesize{Male}} (v9);
\draw[->] (v3) --  node [above, sloped] {\footnotesize{Female}} (v10);
\draw[->] (v4) --  node [below, sloped] {\footnotesize{Male}} (v11);
\draw[->] (v4) --  node [above, sloped] {\footnotesize{Female}} (v12);
\draw[->] (v5) --  node [below, sloped] {\footnotesize{Male}} (v13);
\draw[->] (v5) --  node [above, sloped] {\footnotesize{Female}} (v14);
\draw[->] (v6) --  node [below, sloped] {\footnotesize{Male}} (v15);
\draw[->] (v6) --  node [above, sloped] {\footnotesize{Female}} (v16);
\draw[->] (v7) --  node [below, sloped] {\footnotesize{Male}} (v17);
\draw[->] (v7) --  node [above, sloped] {\footnotesize{Female}} (v18);
\draw[->] (v8) --  node [below, sloped] {\footnotesize{Male}} (v19);
\draw[->] (v8) --  node [above, sloped] {\footnotesize{Female}} (v20);
\draw[->] (v9) --  node [below, sloped] {\footnotesize{No}} (v21);
\draw[->] (v9) --  node [above, sloped] {\footnotesize{Yes}} (v22);
\draw[->] (v10) --  node [below, sloped] {\footnotesize{No}} (v23);
\draw[->] (v10) --  node [above, sloped] {\footnotesize{Yes}} (v24);
\draw[->] (v11) --  node [below, sloped] {\footnotesize{No}} (v25);
\draw[->] (v11) --  node [above, sloped] {\footnotesize{Yes}} (v26);
\draw[->] (v12) --  node [below, sloped] {\footnotesize{No}} (v27);
\draw[->] (v12) --  node [above, sloped] {\footnotesize{Yes}} (v28);
\draw[->] (v13) --  node [below, sloped] {\footnotesize{No}} (v29);
\draw[->] (v13) --  node [above, sloped] {\footnotesize{Yes}} (v30);
\draw[->] (v14) --  node [below, sloped] {\footnotesize{No}} (v31);
\draw[->] (v14) --  node [above, sloped] {\footnotesize{Yes}} (v32);
\draw[->] (v15) --  node [below, sloped] {\footnotesize{No}} (v33);
\draw[->] (v15) --  node [above, sloped] {\footnotesize{Yes}} (v34);
\draw[->] (v16) --  node [below, sloped] {\footnotesize{No}} (v35);
\draw[->] (v16) --  node [above, sloped] {\footnotesize{Yes}} (v36);
\draw[->] (v17) --  node [below, sloped] {\footnotesize{No}} (v37);
\draw[->] (v17) --  node [above, sloped] {\footnotesize{Yes}} (v38);
\draw[->] (v18) --  node [below, sloped] {\footnotesize{No}} (v39);
\draw[->] (v18) --  node [above, sloped] {\footnotesize{Yes}} (v40);
\draw[->] (v19) --  node [below, sloped] {\footnotesize{No}} (v41);
\draw[->] (v19) --  node [above, sloped] {\footnotesize{Yes}} (v42);
\draw[->] (v20) --  node [below, sloped] {\footnotesize{No}} (v43);
\draw[->] (v20) --  node [above, sloped] {\footnotesize{Yes}} (v44);
\draw[->] (v21) --    (v45);
\draw[->] (v21) --    (v46);
\draw[->] (v21) --    (v47);
\draw[->] (v22) --    (v48);
\draw[->] (v22) --   (v49);
\draw[->] (v22) --   (v50);
\draw[->] (v23) --   (v51);
\draw[->] (v23) --  (v52);
\draw[->] (v23) --   (v53);
\draw[->] (v24) --   (v54);
\draw[->] (v24) --  (v55);
\draw[->] (v24) --   (v56);
\draw[->] (v25) --   (v57);
\draw[->] (v25) --  (v58);
\draw[->] (v25) --   (v59);
\draw[->] (v26) --   (v63);
\draw[->] (v26) --  (v61);
\draw[->] (v26) --   (v62);
\draw[->] (v27) --   (v64);
\draw[->] (v27) --  (v65);
\draw[->] (v27) --   (v66);
\draw[->] (v28) --   (v67);
\draw[->] (v28) --  (v68);
\draw[->] (v28) --   (v69);
\draw[->] (v29) --   (v70);
\draw[->] (v29) --  (v71);
\draw[->] (v29) --   (v72);
\draw[->] (v30) --   (v73);
\draw[->] (v30) --  (v74);
\draw[->] (v30) --   (v75);
\draw[->] (v31) --   (v76);
\draw[->] (v31) --  (v77);
\draw[->] (v31) --   (v78);
\draw[->] (v32) --   (v79);
\draw[->] (v32) --  (v80);
\draw[->] (v32) --   (v81);
\draw[->] (v33) --   (v82);
\draw[->] (v33) --  (v83);
\draw[->] (v33) --   (v84);
\draw[->] (v34) --   (v85);
\draw[->] (v34) --  (v86);
\draw[->] (v34) --   (v87);
\draw[->] (v35) --   (v88);
\draw[->] (v35) --  (v89);
\draw[->] (v35) --   (v90);
\draw[->] (v36) --   (v91);
\draw[->] (v36) --  (v92);
\draw[->] (v36) --   (v93);
\draw[->] (v37) --   (v94);
\draw[->] (v37) --  (v95);
\draw[->] (v37) --   (v96);
\draw[->] (v38) --   (v97);
\draw[->] (v38) --  (v98);
\draw[->] (v38) --   (v99);
\draw[->] (v39) --   (v100);
\draw[->] (v39) --  (v101);
\draw[->] (v39) --   (v102);
\draw[->] (v40) --   (v103);
\draw[->] (v40) --  (v104);
\draw[->] (v40) --   (v105);
\draw[->] (v41) --   (v106);
\draw[->] (v41) --  (v107);
\draw[->] (v41) --   (v108);
\draw[->] (v42) --   (v109);
\draw[->] (v42) --  (v110);
\draw[->] (v42) --   (v111);
\draw[->] (v43) --   (v112);
\draw[->] (v43) --  (v113);
\draw[->] (v43) --   (v114);
\draw[->] (v44) --   (v115);
\draw[->] (v44) --  (v116);
\draw[->] (v44) --   (v117);
\end{tikzpicture}
\;\;\;\;\;\;\;\;\;\;\;\;\;\;
\begin{tikzpicture}
\renewcommand{\xx}{2}
\renewcommand{\yy}{1}
\node (v0) at (0*\xx,0*\yy) {\sage{cyan}{0}};
\node (v1) at (1*\xx,-6*\yy) {\sage{cyan}{1}};
\node (v2) at (1*\xx,6*\yy) {\sage{red}{2}};
\node (v3) at (2*\xx,-10*\yy) {\sage{cyan}{3}};
\node (v4) at (2*\xx,-6*\yy) {\sage{yellow}{4}};
\node (v5) at (2*\xx,-2*\yy) {\sage{red}{5}};
\node (v6) at (2*\xx,2*\yy) {\sage{green}{6}};
\node (v7) at (2*\xx,6*\yy) {\sage{green}{7}};
\node (v8) at (2*\xx,10*\yy) {\sage{green}{8}};
\node (v9) at (3*\xx,-11*\yy) {\sage{cyan}{9}};
\node (v10) at (3*\xx,-9*\yy) {\sage{cyan}{10}};
\node (v11) at (3*\xx,-7*\yy) {\sage{red}{11}};
\node (v12) at (3*\xx,-5*\yy) {\sage{red}{12}};
\node (v13) at (3*\xx,-3*\yy) {\sage{green}{13}};
\node (v14) at (3*\xx,-1*\yy) {\sage{green}{14}};
\node (v15) at (3*\xx,1*\yy) {\sage{yellow}{15}};
\node (v16) at (3*\xx,3*\yy) {\sage{yellow}{16}};
\node (v17) at (3*\xx,5*\yy) {\sage{magenta}{17}};
\node (v18) at (3*\xx,7*\yy) {\sage{magenta}{18}};
\node (v19) at (3*\xx,9*\yy) {\sage{green}{19}};
\node (v20) at (3*\xx,11*\yy) {\sage{green}{20}};
\node (v21) at (4*\xx,-11.5*\yy) {\sage{cyan}{21}};
\node (v22) at (4*\xx,-10.5*\yy) {\sage{red}{22}};
\node (v23) at (4*\xx,-9.5*\yy) {\sage{cyan}{23}};
\node (v24) at (4*\xx,-8.5*\yy) {\sage{red}{24}};
\node (v25) at (4*\xx,-7.5*\yy) {\sage{green}{25}};
\node (v26) at (4*\xx,-6.5*\yy) {\sage{yellow}{26}};
\node (v27) at (4*\xx,-5.5*\yy) {\sage{green}{27}};
\node (v28) at (4*\xx,-4.5*\yy) {\sage{yellow}{28}};
\node (v29) at (4*\xx,-3.5*\yy) {\sage{magenta}{29}};
\node (v30) at (4*\xx,-2.5*\yy) {\sage{orange}{30}};
\node (v31) at (4*\xx,-1.5*\yy) {\sage{magenta}{31}};
\node (v32) at (4*\xx,-0.5*\yy) {\sage{magenta}{32}};
\node (v33) at (4*\xx,0.5*\yy) {\sage{gray}{33}};
\node (v34) at (4*\xx,1.5*\yy) {\sage{orange}{34}};
\node (v35) at (4*\xx,2.5*\yy) {\sage{gray}{35}};
\node (v36) at (4*\xx,3.5*\yy) {\sage{orange}{36}};
\node (v37) at (4*\xx,4.5*\yy) {\sage{gray}{37}};
\node (v38) at (4*\xx,5.5*\yy) {\sage{orange}{38}};
\node (v39) at (4*\xx,6.5*\yy) {\sage{gray}{39}};
\node (v40) at (4*\xx,7.5*\yy) {\sage{orange}{40}};
\node (v41) at (4*\xx,8.5*\yy) {\sage{gray}{41}};
\node (v42) at (4*\xx,9.5*\yy) {\sage{orange}{42}};
\node (v43) at (4*\xx,10.5*\yy) {\sage{gray}{43}};
\node (v44) at (4*\xx,11.5*\yy) {\sage{orange}{44}};
\node (v45) at (5*\xx,-11.83*\yy) {\leaf};
\node (v46) at (5*\xx,-11.5*\yy) {\leaf};
\node (v47) at (5*\xx,-11.17*\yy) {\leaf};
\node (v48) at (5*\xx,-10.83*\yy) {\leaf};
\node (v49) at (5*\xx,-10.5*\yy) {\leaf};
\node (v50) at (5*\xx,-10.17*\yy) {\leaf};
\node (v51) at (5*\xx,-9.83*\yy) {\leaf};
\node (v52) at (5*\xx,-9.5*\yy) {\leaf};
\node (v53) at (5*\xx,-9.17*\yy) {\leaf};
\node (v54) at (5*\xx,-8.83*\yy) {\leaf};
\node (v55) at (5*\xx,-8.5*\yy) {\leaf};
\node (v56) at (5*\xx,-8.17*\yy) {\leaf};
\node (v57) at (5*\xx,-7.83*\yy) {\leaf};
\node (v58) at (5*\xx,-7.5*\yy) {\leaf};
\node (v59) at (5*\xx,-7.17*\yy) {\leaf};
\node (v61) at (5*\xx,-6.83*\yy) {\leaf};
\node (v62) at (5*\xx,-6.5*\yy) {\leaf};
\node (v63) at (5*\xx,-6.17*\yy) {\leaf};
\node (v64) at (5*\xx,-5.83*\yy) {\leaf};
\node (v65) at (5*\xx,-5.5*\yy) {\leaf};
\node (v66) at (5*\xx,-5.17*\yy) {\leaf};
\node (v67) at (5*\xx,-4.83*\yy) {\leaf};
\node (v68) at (5*\xx,-4.5*\yy) {\leaf};
\node (v69) at (5*\xx,-4.17*\yy) {\leaf};
\node (v70) at (5*\xx,-3.83*\yy) {\leaf};
\node (v71) at (5*\xx,-3.5*\yy) {\leaf};
\node (v72) at (5*\xx,-3.17*\yy) {\leaf};
\node (v73) at (5*\xx,-2.83*\yy) {\leaf};
\node (v74) at (5*\xx,-2.5*\yy) {\leaf};
\node (v75) at (5*\xx,-2.17*\yy) {\leaf};
\node (v76) at (5*\xx,-1.83*\yy) {\leaf};
\node (v77) at (5*\xx,-1.5*\yy) {\leaf};
\node (v78) at (5*\xx,-1.17*\yy) {\leaf};
\node (v79) at (5*\xx,-0.83*\yy) {\leaf};
\node (v80) at (5*\xx,-0.5*\yy) {\leaf};
\node (v81) at (5*\xx,-0.17*\yy) {\leaf};
\node (v117) at (5*\xx,11.83*\yy) {\leaf};
\node (v116) at (5*\xx,11.5*\yy) {\leaf};
\node (v115) at (5*\xx,11.17*\yy) {\leaf};
\node (v114) at (5*\xx,10.83*\yy) {\leaf};
\node (v113) at (5*\xx,10.5*\yy) {\leaf};
\node (v112) at (5*\xx,10.17*\yy) {\leaf};
\node (v111) at (5*\xx,9.83*\yy) {\leaf};
\node (v110) at (5*\xx,9.5*\yy) {\leaf};
\node (v109) at (5*\xx,9.17*\yy) {\leaf};
\node (v108) at (5*\xx,8.83*\yy) {\leaf};
\node (v107) at (5*\xx,8.5*\yy) {\leaf};
\node (v106) at (5*\xx,8.17*\yy) {\leaf};
\node (v105) at (5*\xx,7.83*\yy) {\leaf};
\node (v104) at (5*\xx,7.5*\yy) {\leaf};
\node (v103) at (5*\xx,7.17*\yy) {\leaf};
\node (v102) at (5*\xx,6.83*\yy) {\leaf};
\node (v101) at (5*\xx,6.5*\yy) {\leaf};
\node (v100) at (5*\xx,6.17*\yy) {\leaf};
\node (v99) at (5*\xx,5.83*\yy) {\leaf};
\node (v98) at (5*\xx,5.5*\yy) {\leaf};
\node (v97) at (5*\xx,5.17*\yy) {\leaf};
\node (v96) at (5*\xx,4.83*\yy) {\leaf};
\node (v95) at (5*\xx,4.5*\yy) {\leaf};
\node (v94) at (5*\xx,4.17*\yy) {\leaf};
\node (v93) at (5*\xx,3.83*\yy) {\leaf};
\node (v92) at (5*\xx,3.5*\yy) {\leaf};
\node (v91) at (5*\xx,3.17*\yy) {\leaf};
\node (v90) at (5*\xx,2.83*\yy) {\leaf};
\node (v89) at (5*\xx,2.5*\yy) {\leaf};
\node (v88) at (5*\xx,2.17*\yy) {\leaf};
\node (v87) at (5*\xx,1.83*\yy) {\leaf};
\node (v86) at (5*\xx,1.5*\yy) {\leaf};
\node (v85) at (5*\xx,1.17*\yy) {\leaf};
\node (v84) at (5*\xx,0.83*\yy) {\leaf};
\node (v83) at (5*\xx,0.5*\yy) {\leaf};
\node (v82) at (5*\xx,0.17*\yy) {\leaf};
\draw[->] (v0) -- node [below, sloped] {\footnotesize{No}} (v1);
\draw[->] (v0) -- node [above, sloped] {\footnotesize{Yes}} (v2);
\draw[->] (v1) --  node [below, sloped] {\footnotesize{No}} (v3);
\draw[->] (v1) --  node [above, sloped] {\footnotesize{Yes}} (v4);
\draw[->] (v1) --  node [above, sloped] {\footnotesize{Unsure}} (v5);
\draw[->] (v2) --  node [below, sloped] {\footnotesize{No}} (v6);
\draw[->] (v2) --  node [above, sloped] {\footnotesize{Yes}} (v7);
\draw[->] (v2) --  node [above, sloped] {\footnotesize{Unsure}} (v8);
\draw[->] (v3) --  node [below, sloped] {\footnotesize{Male}} (v9);
\draw[->] (v3) --  node [above, sloped] {\footnotesize{Female}} (v10);
\draw[->] (v4) --  node [below, sloped] {\footnotesize{Male}} (v11);
\draw[->] (v4) --  node [above, sloped] {\footnotesize{Female}} (v12);
\draw[->] (v5) --  node [below, sloped] {\footnotesize{Male}} (v13);
\draw[->] (v5) --  node [above, sloped] {\footnotesize{Female}} (v14);
\draw[->] (v6) --  node [below, sloped] {\footnotesize{Male}} (v15);
\draw[->] (v6) --  node [above, sloped] {\footnotesize{Female}} (v16);
\draw[->] (v7) --  node [below, sloped] {\footnotesize{Male}} (v17);
\draw[->] (v7) --  node [above, sloped] {\footnotesize{Female}} (v18);
\draw[->] (v8) --  node [below, sloped] {\footnotesize{Male}} (v19);
\draw[->] (v8) --  node [above, sloped] {\footnotesize{Female}} (v20);
\draw[->] (v9) --  node [below, sloped] {\footnotesize{No}} (v21);
\draw[->] (v9) --  node [above, sloped] {\footnotesize{Yes}} (v22);
\draw[->] (v10) --  node [below, sloped] {\footnotesize{No}} (v23);
\draw[->] (v10) --  node [above, sloped] {\footnotesize{Yes}} (v24);
\draw[->] (v11) --  node [below, sloped] {\footnotesize{No}} (v25);
\draw[->] (v11) --  node [above, sloped] {\footnotesize{Yes}} (v26);
\draw[->] (v12) --  node [below, sloped] {\footnotesize{No}} (v27);
\draw[->] (v12) --  node [above, sloped] {\footnotesize{Yes}} (v28);
\draw[->] (v13) --  node [below, sloped] {\footnotesize{No}} (v29);
\draw[->] (v13) --  node [above, sloped] {\footnotesize{Yes}} (v30);
\draw[->] (v14) --  node [below, sloped] {\footnotesize{No}} (v31);
\draw[->] (v14) --  node [above, sloped] {\footnotesize{Yes}} (v32);
\draw[->] (v15) --  node [below, sloped] {\footnotesize{No}} (v33);
\draw[->] (v15) --  node [above, sloped] {\footnotesize{Yes}} (v34);
\draw[->] (v16) --  node [below, sloped] {\footnotesize{No}} (v35);
\draw[->] (v16) --  node [above, sloped] {\footnotesize{Yes}} (v36);
\draw[->] (v17) --  node [below, sloped] {\footnotesize{No}} (v37);
\draw[->] (v17) --  node [above, sloped] {\footnotesize{Yes}} (v38);
\draw[->] (v18) --  node [below, sloped] {\footnotesize{No}} (v39);
\draw[->] (v18) --  node [above, sloped] {\footnotesize{Yes}} (v40);
\draw[->] (v19) --  node [below, sloped] {\footnotesize{No}} (v41);
\draw[->] (v19) --  node [above, sloped] {\footnotesize{Yes}} (v42);
\draw[->] (v20) --  node [below, sloped] {\footnotesize{No}} (v43);
\draw[->] (v20) --  node [above, sloped] {\footnotesize{Yes}} (v44);
\draw[->] (v21) --    (v45);
\draw[->] (v21) --    (v46);
\draw[->] (v21) --    (v47);
\draw[->] (v22) --    (v48);
\draw[->] (v22) --   (v49);
\draw[->] (v22) --   (v50);
\draw[->] (v23) --   (v51);
\draw[->] (v23) --  (v52);
\draw[->] (v23) --   (v53);
\draw[->] (v24) --   (v54);
\draw[->] (v24) --  (v55);
\draw[->] (v24) --   (v56);
\draw[->] (v25) --   (v57);
\draw[->] (v25) --  (v58);
\draw[->] (v25) --   (v59);
\draw[->] (v26) --   (v63);
\draw[->] (v26) --  (v61);
\draw[->] (v26) --   (v62);
\draw[->] (v27) --   (v64);
\draw[->] (v27) --  (v65);
\draw[->] (v27) --   (v66);
\draw[->] (v28) --   (v67);
\draw[->] (v28) --  (v68);
\draw[->] (v28) --   (v69);
\draw[->] (v29) --   (v70);
\draw[->] (v29) --  (v71);
\draw[->] (v29) --   (v72);
\draw[->] (v30) --   (v73);
\draw[->] (v30) --  (v74);
\draw[->] (v30) --   (v75);
\draw[->] (v31) --   (v76);
\draw[->] (v31) --  (v77);
\draw[->] (v31) --   (v78);
\draw[->] (v32) --   (v79);
\draw[->] (v32) --  (v80);
\draw[->] (v32) --   (v81);
\draw[->] (v33) --   (v82);
\draw[->] (v33) --  (v83);
\draw[->] (v33) --   (v84);
\draw[->] (v34) --   (v85);
\draw[->] (v34) --  (v86);
\draw[->] (v34) --   (v87);
\draw[->] (v35) --   (v88);
\draw[->] (v35) --  (v89);
\draw[->] (v35) --   (v90);
\draw[->] (v36) --   (v91);
\draw[->] (v36) --  (v92);
\draw[->] (v36) --   (v93);
\draw[->] (v37) --   (v94);
\draw[->] (v37) --  (v95);
\draw[->] (v37) --   (v96);
\draw[->] (v38) --   (v97);
\draw[->] (v38) --  (v98);
\draw[->] (v38) --   (v99);
\draw[->] (v39) --   (v100);
\draw[->] (v39) --  (v101);
\draw[->] (v39) --   (v102);
\draw[->] (v40) --   (v103);
\draw[->] (v40) --  (v104);
\draw[->] (v40) --   (v105);
\draw[->] (v41) --   (v106);
\draw[->] (v41) --  (v107);
\draw[->] (v41) --   (v108);
\draw[->] (v42) --   (v109);
\draw[->] (v42) --  (v110);
\draw[->] (v42) --   (v111);
\draw[->] (v43) --   (v112);
\draw[->] (v43) --  (v113);
\draw[->] (v43) --   (v114);
\draw[->] (v44) --   (v115);
\draw[->] (v44) --  (v116);
\draw[->] (v44) --   (v117);
\end{tikzpicture}
}
\caption{Staged tree learned using the hill-climbing algorithm (left) and the CSBHC algorithm (right) over the aspects of everyday life data with variables' order S, F, G, B, L.\label{fig:lifehc}}
\end{figure}
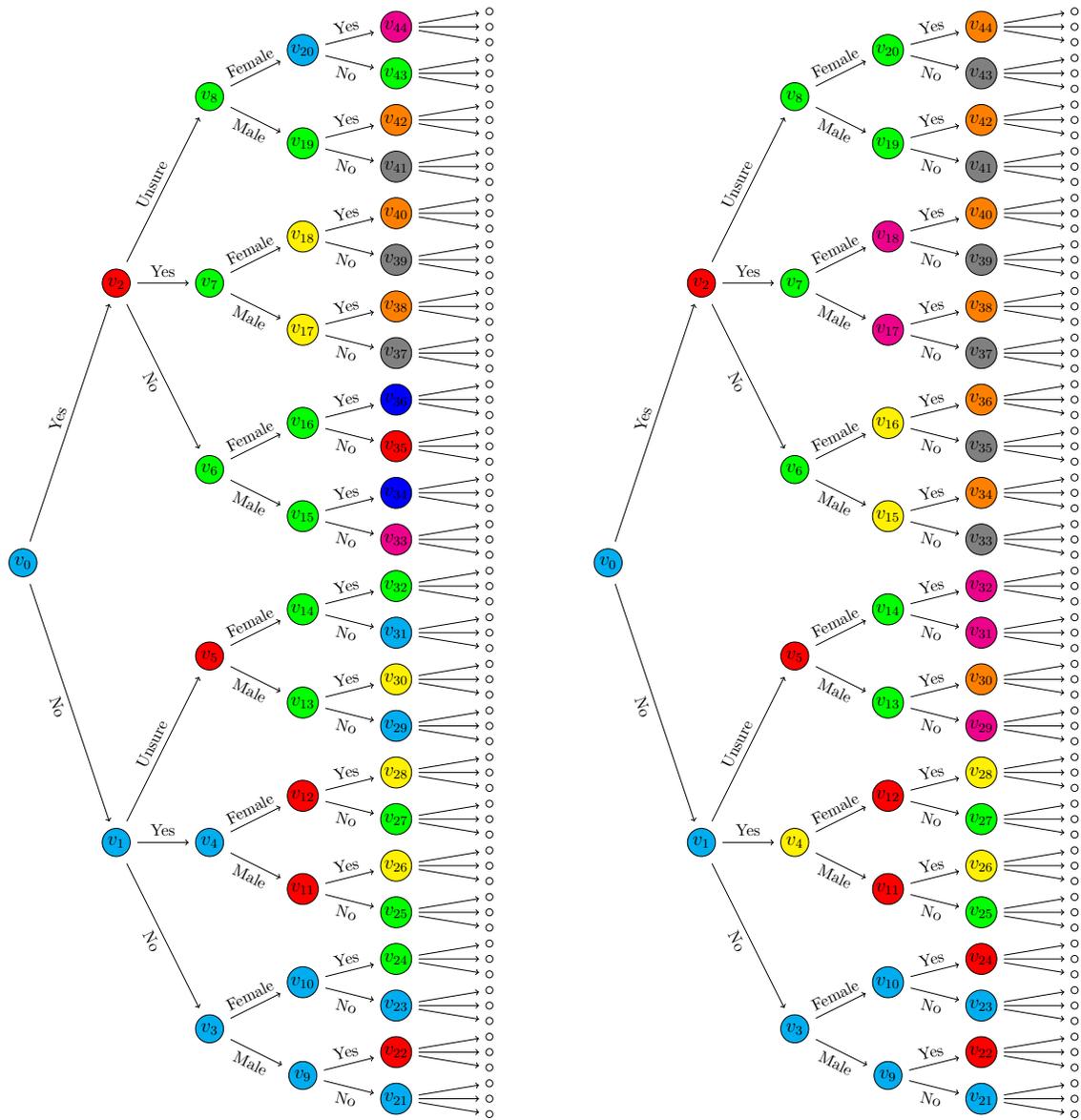

A staged tree over the same dataset is learned as follows. First, we learn a staged tree with the hill-climbing 
algorithm (optimizing BIC score) and 
considering all possible orders of the variables 
but life grade (L), which we then fix as the last variable of the tree. 
The resulting staged tree is plotted in Figure 
\ref{fig:lifehc} (left).

The staging of the life grade variable reveals a complex dependence pattern from which interesting conclusions can be drawn. For instance, conditionally on whether individuals believe in people, life grade does not depend on gender for those that practice sports and have friends they can count on (stages $v_{37}$-$v_{40}$). Similarly, conditionally on whether individuals believe in people, the distribution of life grade is the same for male individuals that practice sports who either have friends or are unsure about it (stages $v_{37},v_{38},v_{41},v_{42}$). It can also be noticed that gender almost has no effect on whether individuals believe in others, the only dependence existing for individuals that practice sports and are unsure about having friends (stages $v_{19}$-$v_{20}$). These are just few of the many conclusions that can be drawn from the tree and a whole explanation is beyond the scope of this analysis.

Although the staged tree in Figure \ref{fig:lifehc} can still be visually inspected, it is already rather extensive with 72 leaves and 45 internal vertices. Its associated ALDAG in Figure \ref{fig:life} (middle) provides a compact summary of the dependence structure and shows that all variables are related to each other according to different types of dependence.

An alternative tree is learned with the CSBHC algorithm 
over the same dataset and variable ordering and is reported in Figure \ref{fig:lifehc} (right). 
The tree has a staging which is a lot more symmetric than the one of the generic staged tree. For instance, it states that life grade is independent of gender and the availability of friends conditionally on whether you believe in people and on conducting sport activity (stages $v_{33}$-$v_{44}$). Also the staging of the vertices $v_{9}$-$v_{20}$ reports that, given a specific level of sports activity and the availability of friends, gender does not affect whether an individual believes in people. The associated ALDAG in Figure \ref{fig:life} (right) is therefore not complete and has a missing edge from G to B. It can also be noticed that edges are only of type total or context. Compared to the standard BN, both ALDAGs show additional patterns of dependence that can be retrieved because the assumption of symmetric dependence was relaxed.

Notice that both the generic staged tree and the alternative staged tree obtained with CSBHC, provide a better description of the dependence structure of the data, since they have lower BICs than the one associated to the BN, 251648 and 251673 respectively. 

\subsection{Enterprise Innovation}

\begin{table}[]
    \centering
    \scalebox{0.75}{
    \begin{tabular}{|c|c|c|}
    \toprule
    Name & Explanation & Levels \\
    \midrule
        GP & Belongs to an industrial group & Yes/No  \\
        LARMAR & Main market & Regional/National/International\\
         INPDDG & Product innovation 2010-2012 & Yes/No \\
         INPDSV & Service innovation 2010-2012 & Yes/No\\
         INPD & Other innovations 2010-2012 & Yes/No \\
         INABA& Abandoned innovation 2008-2010 & Yes/No\\
         INONG & Ongoing innovation from 2008-2010 & Yes/No\\
         CO & Cooperation agreements for innovation & Yes/No\\
         ORG & New organization practices & Yes/No\\
         MKT & New marketing practices & Yes/No\\
         PUB & Contracts with public institutions & Yes/No\\
         EMP & Number of employees & 10-49/50-249/$>$250\\
         EMPUD & Employees with degree & 0\%/1-10\%/$>$10\%\\
         RD &  Research \& development & Yes/No\\
         GROWTH & Increased revenue 2012/2010 & Yes/No\\
         \bottomrule
    \end{tabular}
    }
    \caption{Variables from the 2012 ISTAT enterprise innovation survey.}
    \label{tab:my_label}
\end{table}

\begin{figure}
    \centering
    \includegraphics{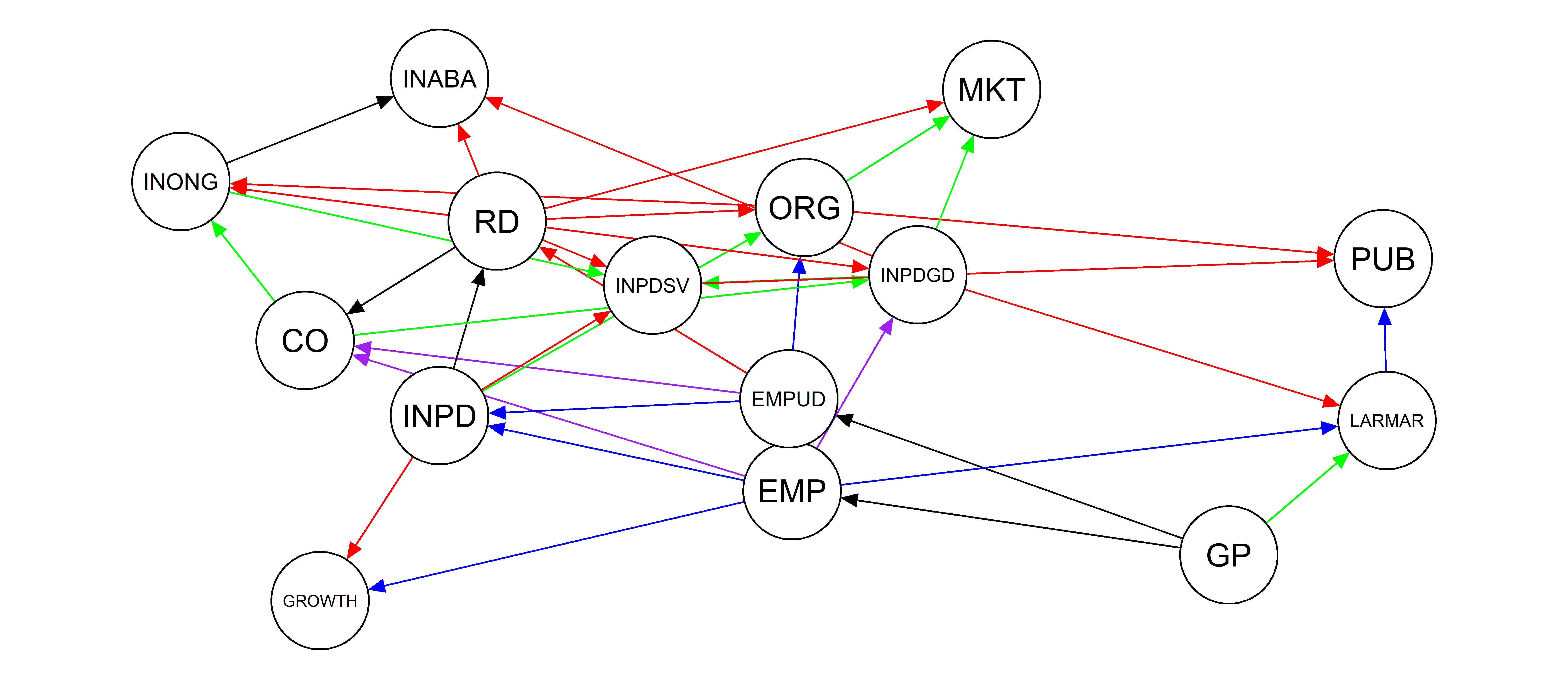}
    \caption{BN (without considering edge coloring) and ALDAG for the enterprise innovation survey data. The edge coloring is: red - context; blue - partial; violet - context/partial; green - local; black - total. }
    \label{fig:innovation}
\end{figure}

We next consider data from the 2012 Italian enterprise innovation survey, again collected by ISTAT \citep{ISTAT2015}. The survey reports information about medium-sized Italian companies as well as their involvement with innovation in the three-years period between 2010-2012. The aim of the analysis is to assess which factors related to innovation are connected with changes in the company revenue. 

Out of the many questions available in the survey, we consider 14 factors that could influence the revenue of an enterprise. The variables considered are summarized in Table \ref{tab:my_label}. Instances with a missing answer were dropped, resulting in 8938 answers to the survey. Notice that in this situation it is unfeasible to study directly the staged tree, since it would have more than 100k leaves making it impossible to visualize its staging.

Therefore, we follow the alternative strategy outlined in Section \ref{sec:learning} of creating an ALDAG as a refinement of a BN. Thus, we first learn a BN from data, reported in Figure \ref{fig:innovation} (by not considering the edge coloring). Interestingly, the BN suggests that the only two factors that have a direct influence on the change of revenue of a company (GROWTH) are the number of employees (EMP) and whether it carried out innovations of other types in the past three years (INPD) - meaning not product or service innovations.

The resulting BN is refined into an ALDAG using the backward hill-climbing algorithm which can only join staged together and is reported in Figure \ref{fig:innovation}. Of the original 35 edges, only five are still of type total and embedding symmetric independences. All other edges are colored, indicating that there are types of dependence in the data which cannot be represented symmetrically. This is confirmed by the BIC of the ALDAG which is equal to 133689, much lower than the one of the BN (134311).

\begin{figure}
\centering
\scalebox{0.8}{
\begin{tikzpicture}
\renewcommand{\xx}{3}
\renewcommand{\yy}{0.6}
\node (v0) at (0*\xx,0*\yy) {\sage{white}{0}};
\node (v1) at (1*\xx,-4*\yy) {\sage{white}{1}};
\node (v2) at (1*\xx,0*\yy) {\sage{white}{2}};
\node (v3) at (1*\xx,4*\yy) {\sage{white}{3}};
\node (v4) at (2*\xx,-5*\yy) {\sage{cyan}{4}};
\node (v5) at (2*\xx,-3*\yy) {\sage{red}{5}};
\node (v6) at (2*\xx,-1*\yy) {\sage{red}{6}};
\node (v7) at (2*\xx,1*\yy) {\sage{green}{7}};
\node (v8) at (2*\xx,3*\yy) {\sage{green}{8}};
\node (v9) at (2*\xx,5*\yy) {\sage{green}{9}};
\node (v10) at (3*\xx,-5.5*\yy) {\leaf};
\node (v11) at (3*\xx,-4.5*\yy) {\leaf};
\node (v12) at (3*\xx,-3.5*\yy) {\leaf};
\node (v13) at (3*\xx,-2.5*\yy) {\leaf};
\node (v14) at (3*\xx,-1.5*\yy) {\leaf};
\node (v15) at (3*\xx,-0.5*\yy) {\leaf};
\node (v16) at (3*\xx,0.5*\yy) {\leaf};
\node (v17) at (3*\xx,1.5*\yy) {\leaf};
\node (v18) at (3*\xx,2.5*\yy) {\leaf};
\node (v19) at (3*\xx,3.5*\yy) {\leaf};
\node (v20) at (3*\xx,4.5*\yy) {\leaf};
\node (v21) at (3*\xx,5.5*\yy) {\leaf};
\draw[->] (v0) -- node [below, sloped] {\footnotesize{10-49}} (v1);
\draw[->] (v0) -- node [below, sloped] {\footnotesize{50-249}} (v2);
\draw[->] (v0) --  node [above, sloped] {\footnotesize{$>$250}} (v3);
\draw[->] (v1) --  node [below, sloped] {\footnotesize{No}} (v4);
\draw[->] (v1) --  node [above, sloped] {\footnotesize{Yes}} (v5);
\draw[->] (v2) --  node [below, sloped] {\footnotesize{No}} (v6);
\draw[->] (v2) --  node [above, sloped] {\footnotesize{Yes}} (v7);
\draw[->] (v3) --  node [below, sloped] {\footnotesize{No}} (v8);
\draw[->] (v3) --  node [above, sloped] {\footnotesize{Yes}} (v9);
\draw[->] (v4) --  node [below, sloped] {\footnotesize{No}} (v10);
\draw[->] (v4) --  node [above, sloped] {\footnotesize{Yes}} (v11);
\draw[->] (v5) --  node [below, sloped] {\footnotesize{No}} (v12);
\draw[->] (v5) --  node [above, sloped] {\footnotesize{Yes}} (v13);
\draw[->] (v6) --  node [below, sloped] {\footnotesize{No}} (v14);
\draw[->] (v6) --  node [above, sloped] {\footnotesize{Yes}} (v15);
\draw[->] (v7) --  node [below, sloped] {\footnotesize{No}} (v16);
\draw[->] (v7) --  node [above, sloped] {\footnotesize{Yes}} (v17);
\draw[->] (v8) --  node [below, sloped] {\footnotesize{No}} (v18);
\draw[->] (v8) --  node [above, sloped] {\footnotesize{Yes}} (v19);
\draw[->] (v9) --  node [below, sloped] {\footnotesize{No}} (v20);
\draw[->] (v9) --  node [above, sloped] {\footnotesize{Yes}} (v21);
\end{tikzpicture}
}

\caption{The dependence subtree associated to the ALDAG in Figure \ref{fig:innovation} for the variable GROWTH. The variable order is (EMP,INPD,GROWTH).  \label{fig:subtree}}
\end{figure}
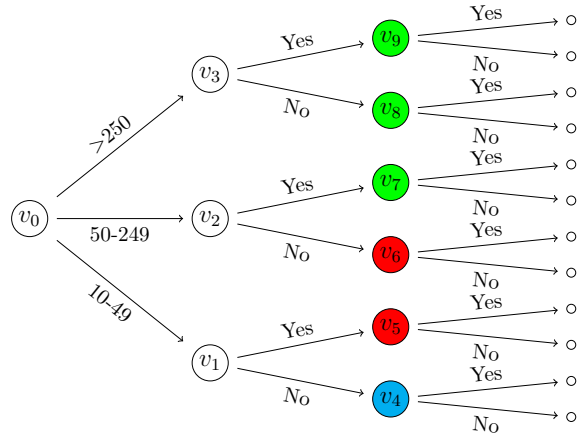

Since GROWTH is independent of all other variables conditionally on EMP and INPD and since our interest is in assessing how factors are relevant for the change of revenue, we can construct a tree with these three variables only and deleting all those that are conditionally independent. We call such a tree the \emph{dependence subtree}. This is reported in Figure \ref{fig:subtree} for GROWTH and the ALDAG in Figure \ref{fig:innovation}. It shows the staging of the variable GROWTH using only its parents in the associated ALDAG. The staging tells us that for larger companies the probability of revenue change does not depend on other innovations and it is the same as for medium-sized companies that invested in other innovations (stages $v_7$-$v_9$).  Medium-sized companies that did not invest in other innovations have the same probability of revenue change as small ones that did invest in other innovations (stages $v_5$-$v_6$). Importantly, larger enterprises and medium-sized ones that invested in other innovations have the larger probability of increasing revenue ($0.61$). On the other hand smaller companies that did not invest in other innovations are more likely to decrease their revenue since their probability of increasing is only $0.47$.

An algorithm for constructing the dependence subtree is based on a simple variation of those given in the supplementary material. Dependence subtrees are extremely powerful since they allow us to visualize the dependence structure of GROWTH by means of the small tree in Figure~\ref{fig:subtree}, without having to investigate the full staged tree having more than 100k leaves.

\section{Discussion}

Staged trees are a flexible class of models that can represent highly non-symmetric relationships. This richness has the drawback that independences are often difficult to assess and visualize intuitively through its graph. In this paper we introduce methods that summarize both the symmetric and non-symmetric relationships learned  from data via structural learning by transforming the tree into a DAG. As a result, we introduced a novel class of graphs extending DAGs by labeling their edges. Our data applications showed the superior fit to data of such models as well as the information they can provide in real domains. 

The new DAG edge labeling is based on the identification of the class of dependence. A different possibility would be to define a dependence index between any two variables which measures how different their relationship is from total dependence/independence. By learning a staged tree from data we could then label the edges of a BN with such indexes. The definition of such models is the focus of current research. 

This work further provides a first criterion to read any symmetric conditional independence from a staged tree. Algorithms to assess if generic non-symmetric conditional independence statements hold still need to be developed. Here we have provided an intermediate solution to this problem by characterizing if a non-symmetric independence exists or not. We plan to provide a conclusive solution to non-symmetric independence queries in future work.

\bibliographystyle{chicago}

\bibliography{refs}

\newpage

\appendix

\section{Proof of Proposition \ref{theo:1}} 
\label{appendixa}
For the proof we use the following Lemma.

\begin{lemma}
	\label{prop:inclusion}
	Let $T = (V, E, \theta)$ and $T' = (V, E, \theta')$ be two staged trees 
	with same tree $(V,E)$.  
	We have that $\mathcal{M}_T \subseteq \mathcal{M}_{T'}$ 
	if and only if for each $e,f \in E$, 
	$\theta'(e) = \theta'(f) \Rightarrow \theta(e) = \theta(f)$. 
\end{lemma} 

The result follows directly from the definition of $\mathcal{M}_T$.

The proof of Proposition \ref{theo:1} starts by showing that $\mathcal{M}_T \subseteq \mathcal{M}_{G_T}$.
	We have that $\mathcal{M}_{G_T}$ = $\mathcal{M}_{T_{G_T}}$ and
	let $\theta'$ be the labeling of $T_{G_T}$.  
	Both $T$ and $T_{G_T}$ are $\bf X$-compatible staged tree and 
	thus share the same vertex and edge set. 
	Suppose that for $\mathbf{x}_{[i-1]}, \mathbf{y}_{[i-1]} \in 
	\mathbb{X}_{[i-1]}$, 
	$\kappa(\mathbf{x}_{[i-1]}) \neq \kappa(\mathbf{y}_{[i-1]})$ . 
	We define now a sequence of nodes 
	$\mathbf{x}_{[ii-1]} = \mathbf{x}^0_{[i-1]}, \ldots, 
	\mathbf{x}_{[i-1]}^{i-1} = \mathbf{y}_{[i-1]}$ in 
	$\mathbb{X}_{[i-1]} \subseteq V$ 
	as following,
	\[ \mathbf{x}^{0}_{[i-1]} = \mathbf{x}_{[i-1]}, 
	\quad x^{h}_{j} = \left\{ 
	\begin{array}{ll}
		x^{h-1}_j & j > h \\
                y_j & j \leq h 
	\end{array}
	\right. . \] 
	Define  $h_0 = \min \{h \text{ s.t. } \kappa(\mathbf{x}^h_{[i-1]}) \neq 
	\kappa(\mathbf{x}^{h-1}_{[i-1]}) \}$. 
	We have that $k(\mathbf{x}_{[i-1]}) = \ldots = 
	\kappa(\mathbf{x}^{h_0-1}_{[i-1]}) \neq 
	\kappa(\mathbf{x}^{h_0}_{[i-1]}) $
	and thus, by construction of $G_T$,  
	$(h_0, i) \in F_T$ that in turn implies $\theta'(\mathbf{x}_{[i-1]}, x_i ) 
	\neq 
	\theta'( \mathbf{y}_{[i-1]}, x_i)$ for all $x_i \in \mathbb{X}_i$.
	We thus have 
	$\mathcal{M}_T \subseteq \mathcal{M}_{T_{G_T}} = \mathcal{M}_{G_T}$
	by Lemma~\ref{prop:inclusion}.  
	
	Assume now $G = ([p], F)$ is a DAG with
        $1,\ldots,p$ as a topological ordering and such that  
	$\mathcal{M}_T 
	\subseteq \mathcal{M}_G$, then  
	$\mathcal{M}_T \subseteq \mathcal{M}_{T_G}$  
	and if $\gamma$ is the labeling of $T_G$ we have that
	$\gamma(e) = 
	\gamma(f) \Rightarrow 
	\theta(e) = 
	\theta(f)$ 
	for each $e,f \in E$  
	by Lemma~\ref{prop:inclusion}. 
	Let $(k,i) \in F_T$ be an edge of $G_T$, then by construction 
	there exist $\mathbf{x}_{[i-1]},\mathbf{x}'_{[i-1]} \in 
	\mathbb{X}_{[i-1]}$ with $x_j = x'_j, j \neq k$ such that 
	$\kappa(\mathbf{x}_{[i-1]}) \neq \kappa(\mathbf{x}'_{[i-1]})$.  
	Hence, $\gamma(\mathbf{x}_{[i-1]}, 
	(\mathbf{x}_{[i-1]}, x_i) ) \neq 
	\gamma(\mathbf{x}'_{[i-1]}, (\mathbf{x}'_{[i-1]}, x_i) ))$ and it is easy 
	to see that 
	this implies 
	$(h,i) \in F$ by construction of $T_G$. 


\section{Algorithms' Implementations} 
\label{appendixb}

The implemented algorithms work with $\bm{X}$-compatible  staged
trees $T = (V, E, \theta)$. 
There are different ways  to practically implement staged trees, here we 
use a representation similar to the R package \texttt{stagedtrees}~\citep{Carli2020}.
In particular, let $\bm{X}$ be a random vector taking values in 
$\mathbb{X} = \times_{i\in [p]} \mathbb{X}_i$ and assume 
there are total orders over each $\mathbb{X}_i$. 
Then  
$\mathbb{X}_{[i]}$ has 
an induced lexicographic total order. 
The node labeling (or coloring) $\kappa$ in the $\bm{X}$-compatible
staged tree definition can be represented by a sequence of vectors of symbols 
$\bm{s}^1, \bm{s}^2, \ldots, \bm{s}^{p-1}$ 
where $\bm{s}^j =\left(\kappa(v)\right)_{v \in \mathbb{X}_{[j]}} \in \mathcal{C}^{|\mathbb{X}_{[j+1]}|}$ is 
the vector of coloring  
and $\mathbb{X}_{[j+1]}$ is the set of nodes of $T$ with depth $j$, ordered with respect to the 
induced lexicographic order. 
The vectors $\bm{s}^1, \ldots, \bm{s}^{p-1}$ represent the 
stages of the tree and thus we refer to them as \textit{stage vectors}. 





Algorithm~\ref{alg:bn_to_staged} describes the pseudo-code for the conversion 
algorithm that takes as input a DAG over $[p]$ and outputs an $\bm{X}$-compatible staged tree. We assume that $1, \ldots, p$ is a topological ordering of the 
DAG nodes, if not a simple permutation is sufficient before applying the algorithm. 

\begin{algorithm}[tb]
\caption{DAG to $\bm{X}$-Compatible  Staged Tree} \label{alg:bn_to_staged}
  \algsetup{linenosize=\tiny}
  \scriptsize
	\begin{algorithmic}
		\REQUIRE A DAG $G = ([p], F)$ such that $1,\ldots,p$ is a
		topological order of $G$.
	\ENSURE The stage vectors $\bm{s}^1, 
		\ldots, \bm{s}^{p-1}$ encoding an  
		$\bf X$-compatible
		staged tree $T$ such that $\mathcal{M}_T = \mathcal{M}_G$.  

	\FOR{$i = 1$ to $p-1$}       
		\STATE  $\bm{s}^i  = [ 1 ]$ 
		\FOR{$j = 1$ to $i$} 
		    \IF{$j \in \Pi_{i+1}$}
		    \STATE  $\bm{s}^i = [ s \times \mathbb{X}_{j} : s 
		    \in \bm{s}^i]$   
		    \ELSE
		    \STATE $\bm{s}^i = [s \times [1, \ldots, 1] : s 
		    \in \bm{s}^i]$  
		    \ENDIF
		\ENDFOR
      	\ENDFOR
	\end{algorithmic}
\end{algorithm}

Algorithm~\ref{alg:staged_to_bn} is an implementation of the 
conversion from staged trees to DAGs. The algorithm records additional information
about the type of dependence between variables, so that it can be used to define the corresponding ALDAG. 

Algorithm~\ref{alg:csbhc} is the pseudo-code of the 
context specific backward hill-climbing algorithm 
described in Section~\ref{sec:csbhc}. 

In the pseudo-code for Algorithms~\ref{alg:staged_to_bn} and~\ref{alg:csbhc} we use the 
following notation: $\operatorname{vec}(A)$ is the column-wise vectorization 
of a matrix $A$ and $\operatorname{mat}^{m,n}(\bm{a})$ is the column-wise $(m,n)$-matrix-filling such that $\operatorname{vec}(\operatorname{mat}^{m,n}(a)) = a$ and $\operatorname{mat}^{m,n}(\operatorname{vec}(A)) = A$, where $a$ is a vector of symbols of length $mn$ and $A$ is a matrix of symbols 
of dimensions $(m,n)$. 

Both Algorithms~\ref{alg:staged_to_bn} and~\ref{alg:csbhc} are based on the following observation.
For every $i \in [p-1]$ consider the following $i$ matrices of stage symbols obtained iteratively:
\begin{align*}
A_{i} &=  \operatorname{mat}^{m_i,n_i}(\mathbf{s}^i)  &
\text{where } m_i = |\mathbb{X}_{i}|, \, 
n_i = \operatorname{lenght}(\mathbf{s}^i) / m_i \\
A^{j} &= \operatorname{mat}^{m_{j},n_{j}}(\operatorname{vec}(A_{j+1}^t))  &
\text{where } m_{j} = |\mathbb{X}_{j}|, \, 
n_j = \operatorname{lenght}(\mathbf{s}^i) / m_j
\end{align*} 
Then we have that for each $j \leq i$ we can easily
read the context, partial and local conditional 
independences between $X_{i+1}$ and $X_j$ from the matrix of stages $A_j$. 
Each row in the matrix $A_j$ 
corresponds to a level of the variable $X_j$ 
and each column corresponds to a 
context of the type $x_{[i] \setminus {j}}$.
Thus, for example, if a column of $A_j$ has all 
elements equal, it means that, for the particular context, the value of $X_j$ does not affect the 
conditional probability for $X_{i+1}$; hence 
$X_{i+1}$ is conditional independent of $X_j$
given the specific context. If a subset of a column has all equal elements, the conditional independence is partial. And finally, if some 
elements of the matrix $A_j$ are equal and belong to different columns then the independence is
only local.

\begin{algorithm}[!h]
\caption{$\bm{X}$-Compatible Staged Tree to DAG} \label{alg:staged_to_bn}
  \algsetup{linenosize=\tiny}
  \scriptsize
	\begin{algorithmic}
	\REQUIRE An $\bf X$-compatible staged tree $T$ encoded 
		with stage vectors $\bm{s}^1, \ldots, \bm{s}^{p-1}$.
		\ENSURE A minimal DAG $G = ([p], F)$ 
		such that $\mathcal{M}_T \subseteq 
		\mathcal{M}_G$ and a labeling of its edges $\psi$ defining 
		an ALDAG.
		\STATE $G = (V, E = \emptyset) $ 
	\FOR{$i = 1$ to $p-1$}       
		\STATE $a = \bm{s}^i$
	\FOR{$j = i$ to $1$} 
	    \STATE $N = \operatorname{length}(a)$
		\STATE $m = |\mathbb{X}_j|$ and $n = N / m$ 
		\STATE $A = \operatorname{mat}^{m,n}(a)$ 
		\STATE $c_k = |\{ A_{u,k} : u \in [m] \}|, \, k \in [n]$
		\STATE $r_u = |\{ A_{u,k} : k \in [n] \}|, \, u \in [m]$
		\IF{$\max \{ c_k : k \in [n]  \} > 1$}
		    \STATE \# $j$ is a parent of $i+1$
		    \STATE $E = E \cup \{ (j, \, i + 1) \}$ 
		    \STATE \# check the type of dependence 
		    \IF{$\min \{ c_k : k \in [n]  \} = m$}  
		       \STATE $r = \sum_{i \in [m]} |\{ A_{u,k} : k \in [n] \}| $ 
		       \STATE $d =  |\{ A_{u,k} : u \in [m], \, k \in [n] \}|$
		       \IF{$r  \neq  d $}
		          \STATE $\psi(j, i+1) =  \text{local}$
		       \ENDIF
		     \ELSIF{$\min \{ c_k : k \in [n]  \} = 1$}  
		        \STATE \# there is at least one context specific indep. 
		        \IF{$\exists k \in [n] \text{ s.t. } 2 < c_k < m$ }
		           \STATE \# there is also a partial indep.
		           \STATE $\psi(j, i+1) = \text{context/partial}$
		        \ELSE
		          \STATE $\psi(j, i+1) = \text{context}$
		        \ENDIF 
		      \ELSE
		         \STATE $\psi(j, i+1) = \text{partial}$ 
		    \ENDIF
		    \STATE $a = \operatorname{vec}(A^t)$
		  \ELSE 
		      \STATE \# all rows of $A$ are equal,
		       and $j$ is not a parent of $i+1$ 
		      \STATE $a = (A_{1,1}, \ldots, A_{1,n})$
		\ENDIF
        \ENDFOR
        \ENDFOR
	\end{algorithmic}

\end{algorithm}

\begin{algorithm}[!h]
\caption{Context-specific backward hill-climb (CSBHC)} \label{alg:csbhc}
  \algsetup{linenosize=\tiny}
  \scriptsize
	\begin{algorithmic}
	\REQUIRE A starting $\bf X$-compatible staged tree $T$ encoded 
		with stage vectors $\bm{s}^1, \ldots, \bm{s}^{p-1}$; 
		a dataset $\mathcal{D}$ of observations from $\bf X$;
		The score $f: (T, \mathcal{D}) \mapsto \mathbb{R}$ to be optimized. 
		\ENSURE An $\bf X$-compatible staged tree 
    \STATE $f_{best} = f(T, D)$, compute initial score
	\FOR{$i = 1$ to $p-1$}
	\REPEAT
		\STATE $a = \bm{s}^i$ the stage vector of $T$ for variable $X_{i+1}$
		\STATE $N = \operatorname{length}(a) = 
		\prod_{j \in [i]} |\mathbb{X}_j|$ 
	\FOR{$j = i$ to $1$} 
		\STATE $m = |\mathbb{X}_j|$ and $n = N / m$ 
		\STATE $A = \operatorname{mat}^{m,n}(a)$
		\FOR{$\alpha$ column of $A$}
		 \STATE let $T'$ be the staged tree 
		 obtained from $T$ by joining together the stages in $\alpha$
		 \IF{$f(T', D) > f_{best}$}
		 \STATE $T_{best} =T'$
		 \STATE $f_{best} = f(T', D)$
		 \ENDIF
		\ENDFOR
		\STATE $a = \operatorname{vec}(A^t)$
    \ENDFOR
    \STATE $T = T_{best}$
    \UNTIL{no improvement in score is possible}
    \ENDFOR
	\end{algorithmic}

\end{algorithm}

\end{document}